\documentclass[10pt,twocolumn,letterpaper]{article}

\usepackage[pagenumbers]{cvpr}              

\usepackage[dvipsnames]{xcolor}

\definecolor{cvprblue}{rgb}{0.21,0.49,0.74}
\usepackage[pagebackref,breaklinks,colorlinks,citecolor=cvprblue]{hyperref}

\usepackage{latexsym}
\usepackage{dirtytalk}
\usepackage{amsmath}
\usepackage{booktabs}
\usepackage{graphicx}
\usepackage{lipsum, color}
\usepackage{mathtools, cuted}
\usepackage{multirow}
\usepackage{array}
\usepackage{dirtytalk}
\usepackage{multicol}
\usepackage{url}
\makeatletter
\g@addto@macro{\UrlBreaks}{\UrlOrds}
\makeatother
\usepackage{array}
\usepackage{graphbox}
\newcolumntype{L}[1]{>{\raggedright\let\newline\\\arraybackslash\hspace{0pt}}m{#1}}
\newcolumntype{C}[1]{>{\centering\let\newline\\\arraybackslash\hspace{0pt}}m{#1}}
\newcolumntype{R}[1]{>{\raggedleft\let\newline\\\arraybackslash\hspace{0pt}}m{#1}}
\usepackage{pifont}
\usepackage{float}
\usepackage{amsmath,amsfonts,amssymb,amsthm}
\usepackage{mathtools}
\usepackage{commath}
\usepackage{caption, subcaption}
\usepackage{breakcites}
\usepackage{xargs}                      

\newcommand{\ignore}[1]{}

\usepackage[inkscapearea=page]{svg}

\usepackage{colortbl}
\usepackage{booktabs}

\newcommand{\longvivit}{\textsc{LongViViT}\xspace}
\newcommand{\shortvivit}{\textsc{ShortViViT}\xspace}
\newcommand{\imagevit}{\textsc{ImageViT}\xspace}

\usepackage[export]{adjustbox}

\def\paperID{15920} 
\def\confName{CVPR}
\def\confYear{2024}

\title{A Simple Recipe for Contrastively Pre-training Video-First Encoders\\Beyond 16 Frames}

\author{Pinelopi Papalampidi\thanks{Equal contribution.} \quad \quad \ Skanda Koppula\footnotemark[1]  \quad \quad \ Shreya Pathak\footnotemark[1]  \\ Justin Chiu  \quad \quad Joe Heyward \quad \quad Viorica Patraucean  \quad \quad Jiajun Shen  \quad \quad Antoine Miech \\ \quad \quad Andrew Zisserman  \quad \quad Aida Nematzadeh\\
Google DeepMind\\
{\tt\small \{pinelopi,skandak,shreyapa\}@google.com}
}

\begin{document}
\maketitle
\begin{abstract}
Understanding long, real-world videos requires modeling of long-range visual dependencies. To this end, we explore video-first architectures, building on the common paradigm of transferring large-scale, image--text models to video via shallow temporal fusion.
However, we expose two limitations to the approach: (1) decreased spatial capabilities, likely due to poor video--language alignment in standard video datasets, and (2) higher memory consumption, bottlenecking the number of frames that can be processed. To mitigate the memory bottleneck, we systematically analyze the memory/accuracy trade-off of various efficient methods: factorized attention, parameter-efficient image-to-video adaptation, input masking, and multi-resolution patchification. Surprisingly, simply masking large  portions of the video (up to 75\%) during contrastive pre-training proves to be one of the most robust ways to scale encoders to videos up to 4.3 minutes at 1 FPS. Our simple approach for training long video-to-text models, which scales to 1B parameters, does not add new architectural complexity and is able to outperform the popular paradigm of using much larger LLMs as an information aggregator over segment-based information on benchmarks with long-range temporal dependencies (YouCook2, EgoSchema).

\end{abstract}    

\begin{figure}[t]
    \tiny
    \centering
    \includegraphics[width=0.9\columnwidth]{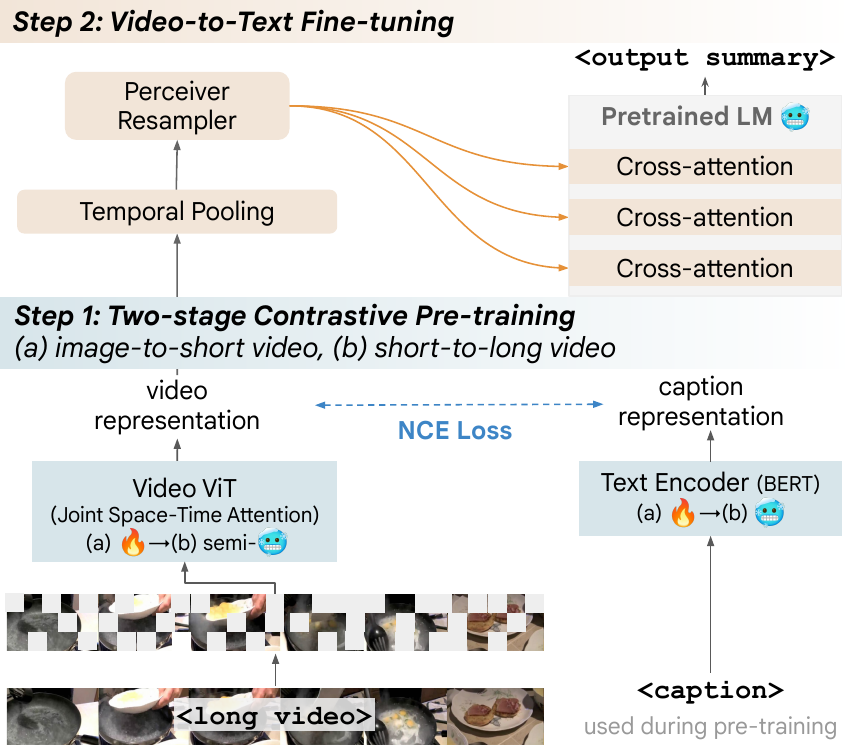}
    \vspace{-1em}
    \caption[ ]
    {\small Two main training steps: (1) training a video encoder via Noise Contrastive Estimation and (2) using this frozen video encoder with a pre-trained, frozen LM and visual adapter layers for video-to-text generation (e.g., video summarization and Q/A).}
    \vspace{-3em}
    \label{fig:model}
\end{figure}

\section{Introduction}
\label{sec:intro}

Long-video understanding requires modeling of the temporal dynamics and long-range visual dependencies of real-world scenes~\cite{wu2022memvit,longvideo}. 
However, capturing long-range visual content is challenging, even when equipped with large language models. In this paper, we overcome hardware memory limitations and demonstrate how to extend video encoders to directly process minutes-long visual content using language grounding, and simple, established techniques without additional architectural complexity~\cite{wu2022memvit, ssm}. We focus on long videos through the lens of language, assessing our models on the widely applicable tasks of visual summarization and question-answering.

Recent work on vision--language models have yielded impressive results, predominantly focusing on understanding images or short clips of 16 frames or less~\cite{alayrac2022flamingo,li2023blip, yang2022zero,ye2023mplug,driess2023palm}. This work recycles strong pre-trained image encoders, performs late temporal fusion~\cite{alayrac2022flamingo,yan2022video,ye2023mplug}, and employs mostly-frozen, powerful LLMs. The lack of \textit{video-first} encoders, equipped with early temporal aggregation, may handicap the ability to process complex visual dependencies, and this is usually reflected in prior work's focus on short video benchmarks ($<30$ seconds) in which sixteen frames are sufficient for competitive performance~\cite{buch2022revisiting,lei2022revealing}.

In this work, we systematically explore video-first models starting from a standard image--language recipe using two-step training and large pre-trained LMs~(Figure~\ref{fig:model};~\cite{alayrac2022flamingo}). This baseline enables us to start from a demonstrably scalable, simpler-to-tune, widely-used recipe that performs competitively~\citep{fu2021violet, li2023blip}.
Through our analysis, we are able to scale this method in a memory-efficient manner to longer sequences of frames, up to 4.3 minutes of video at 1 FPS. 

We first explore video-first models on short-video benchmarks (MSR-VTT~\cite{xu2016msr}, VATEX~\cite{wang2019vatex}, YouCook2~\cite{zhou2018towards}, ActivityNet~\cite{krishna2017dense}) and compare against the SoTA VideoCoCa model~\citep{yan2022video}. We show that simple joint space-time attention significantly improves performance over frame-level encodings on benchmarks with rich temporal dependencies (YouCook2, VATEX).
Overall, our models are able to reach VideoCoCa performance, while requiring fewer parameters and lower frame resolution.

This performance gain incurs extra compute and memory costs that grow quadratically with the video length. To address this, we provide one of the first systematic analyses of the memory/accuracy pareto-front of popular memory-efficient methods; this includes factorized attention, parameter-efficient image-to-video adaptation, input masking, and multi-resolution patchification.
Through this analysis, we find that among all these options, simple token masking (up to 75\%) during contrastive pre-training incurs only a 1\% Recall@1 drop on zero-shot text-video retrieval, and no drop in zero-shot video captioning.
At the same time, such high masking offers 2-3x memory savings and allows us to generalize to longer video contexts. The alternatives we explore (\eg, efficient backbone architectures, more sophisticated TubeViT-style patchification~\cite{piergiovanni2023rethinking}), do not maintain the same robustness against noisy video inputs and present a 25\% relative decrease in performance for text-video retrieval on challenging benchmarks (YouCook2, VATEX). Finally, although parameter-efficient methods~\cite{houlsby2019parameter,hulora} fail to adapt image encoders to video-first models without suffering performance drops, we find that they can adapt video models trained on short contexts (e.g., 16 second videos) to longer temporal horizons.

Based on the above learnings, we extend our best performing short-video encoder to longer contexts of 256 frames (4.3 minutes at 1 FPS). We use the full-length videos of HowTo100M~\cite{miech2019howto100m} accompanied by LLM-generated summaries based on the ASR to further contrastively train our \longvivit while masking 75\% of the input video tokens and freezing most parameters of the encoder. \longvivit-to-text ($\sim$1B parameters) is able to outperform modular methods that use LLM assistance and PALI-3~\cite{chen2023pali} for frame captioning on temporally rich benchmarks (YouCook2, EgoSchema). Even modular methods that consider frame selection (SeViLA~\cite{yu2023self}) or an oracle segmentation of the video for localizing and captioning key events (on YouCook2) cannot reach \longvivit's performance. An interesting byproduct of our work is that we can glean which video--language benchmarks have strong temporal dependencies, and thus are suitable for testing long video models; we find that papers often use benchmarks in which short video or even blind models perform well~\cite{xu2016msr, caba2015activitynet,mangalam2023egoschema}.

\noindent In short, we provide the following contributions:
\begin{itemize}
    \item We  explore the memory/accuracy pareto-frontier of video-first vision--language models, and systematically evaluate many architectural, data, and training alternatives. In the end, we identify a simple recipe that enables scaling to 4.3 minutes at 1 FPS, many times longer than comparable video--language models~\cite{yan2022video, alayrac2022flamingo}.
    \item We identify short and long video benchmarks with substantial temporal dependencies, for which we demonstrate that the traditional image-first, late-temporal fusion recipe is convincingly weaker than a video-first approach.
    \item Finally, we compare our long video models to a variety of strong baselines and show competitive performance with far fewer parameters; this includes baselines that use LLM-based aggregation over visual captions, and we quantitatively evaluate this common approach for the first time on standard video benchmarks.
\end{itemize}

\vspace{-0.5em}
\section{Related Work}

We base our recipes on~\cite{alayrac2022flamingo, li2023blip}, which provide a strong two-step video--language recipe that leverages pre-trained LLMs and works at scale. Similar work at smaller scale has additionally included captioning losses~\cite{zellers2021merlot,li2023lavender}, more contrastive losses~\cite{miech2020end,xu2021videoclip,luo2022clip4clip,cheng2023vindlu}, masking/masked autoencoding~\cite{flip,he2021masked, tongvideomae, fu2021violet,fu2023empirical,lin2023smaug,ma2022simvtp, han2022turbo}, and combinations thereof~\cite{yang2022zero, wang2022internvideo,singh2022flava,wang2022omnivl,yuan2021florence, ye2023mplug,huang2023language,driess2023palm,zhu2023minigpt}. This work focuses on image--text modeling and extends to $<$30 seconds via image-to-video transfer, selective fine-tuning, or temporal fusion of frame encodings~\cite{alayrac2022flamingo,yan2022video, ye2023mplug}.

A volume of work focuses on video-first learning. This includes some of the very early work in image-to-video kernel inflation~\cite{carreira2017quo,simonyan2014two,tran2018closer}, transformer-based video architectures~\cite{arnab2021vivit,bertasius2021timesformer,liu2022video}, image-to-video parameter-efficient adaption~\cite{pan2022st,liu2023revisiting,chen2022litevl}, and multiple spatiotemporal resolutions along different network paths~\cite{ma2022rethinking, feichtenhofer2019slowfast, yan2022multiview,xue2022advancing}. These have still only been demonstrated on short videos, so other works have broached the challenge of temporal scalability:~\cite{wu2022memvit, ssm, ryoo2023token} propose alternative encoders, and~\cite{peng2023yarn, wang2020linformer, kitaev2020reformer} propose more exotic attention mechanisms. TubeViT~\cite{piergiovanni2023rethinking} proposes multi-granularity patchification. We systematically dissect what works and scales among some of these alternatives, electing options that enable us to re-use strong pre-trained models and use standard, more easily-tuned architectures.

Specifically in video-to-text generation, approaches that handle longer videos are very limited and mostly target images or short videos~\cite{wang2022internvideo, li2023videochat, fu2021violet}. A dominant approach is to summarize frames and aggregate information via LLMs~\cite{zeng2022socratic,wang2022language,li2023videochat,lin2023mmvid}. To the best of our knowledge, we are the first to attempt to train large-scale video-to-text models on longer sequences of frames and directly test them against LLM-assisted modular methods on challenging temporal  benchmarks~\cite{zhou2018towards,mangalam2023egoschema}.
\section{The Video-to-Text Architecture} 

We base our approach on the successful two-step recipe that combines pre-trained vision and language models \citep[\eg,][]{alayrac2022flamingo,li2023blip, yang2022zero,ye2023mplug} as shown in Figure~\ref{fig:model}: (1) we first pre-train a vision encoder, and then (2) fuse the frozen vision representations into a pre-trained, frozen LM.

\subsection{Video--Language Contrastive Pre-training} \label{sec:contrastive_pretraining}

Following common practice~\citep[][]{alayrac2022flamingo, li2023blip}, we use a dual vision--language architecture with a Noise Contrastive Estimation (NCE) loss~\cite{gutmann2010noise,wu2018unsupervised,oord2018representation} to pre-train our vision encoder, similar to CLIP~\cite{radford2021learning}, ALIGN~\cite{jia2021scaling} and VideoCLIP~\cite{xu2021videoclip}. 
Both encoders are transformers~\cite{vaswani2017attention}: a BERT-medium (77M) or base (117M) language encoder and ViT-Base (86M parameters) or Large (307M parameters) vision encoder. On the language side, caption representations are computed by averaging across the corresponding token representations. On the vision side, video frames are patchified into a sequence of visual tokens, fed into a vision encoder, and then average pooled to produce a final video representation.

Most prior larger-scale video--language models use pre-trained image encoders and patchify frames individually via 2D convolutions \citep[\eg,][]{yan2022video,xu2021videoclip,alayrac2022flamingo}. Instead, we create spatiotemporal tubelets via 3D convolutions as done in recent vision-only models \citep[][]{arnab2021vivit,tongvideomae,piergiovanni2023rethinking}. Using 3D tubelets instead of flat patches has the dual advantage of higher input compression and more explicit temporal contextualization; our early experiments yielded improved performance. The tubelet embedding sequence is then flattened, added to learnable positional embeddings, and fed into the vision encoder. The vision encoder uses spatio-temporal attention as in ViViT~\citep[][]{arnab2021vivit}: \textit{Joint space-time attention} does not add any new parameters to vanilla image ViT \citep[][]{dosovitskiy2020image}, facilitating transfer between image and video models.

Training a large-scale transformer-based video encoder can be challenging because self-attention across thousands of visual tokens is both compute and memory intensive. Memory bottlenecks a model in two ways: (1) limiting the number of frames, and (2) limiting the contrastive batch size during training, negatively impacting performance. To address (2), we use a pre-trained image encoder trained with large batch sizes, and further tune it on videos, instead of jointly training from scratch on images and videos. For initializing the 3D convolution, we repeat the pre-trained weights across the temporal dimension similarly to~\cite{arnab2021vivit} (see Appendix A). During video--language pre-training, we maintain different embedding paths for images vs. videos: images are embedded with the original 2D convolution and videos with a separate 3D convolution (no weight sharing).

\subsection{Video-to-Text Tuning} \label{sec:video_to_text}

We follow prior work \citep[\eg,][]{alayrac2022flamingo,yang2022zero,ye2023mplug} by plugging the frozen pre-trained vision encoder into a frozen pre-trained LM. We first temporally mean pool the video representations to keep a fixed number of tokens independently of the number of frames and next use a randomly initialized Perceiver-resampler~\cite{jaegle2021perceiver} to project the representations to the LM embedding space (Appendix A). We add new randomly initialized cross-attention layers at each layer of the LM to ground generation on the visual content. We train the new layers and Perceiver resampler with a standard auto-regressive video captioning loss: $- \log p(w_t|w<t;\mathcal{V})$,
where $w_t$ is its $t^{th}$ token, and $\mathcal{V}$ is the video representation. 

\section{Memory-Efficient Encoder Design Space} 
\label{sec:methods_design_space}

Device memory is a key bottleneck for video training with joint space-time attention. To overcome this, we explore four broad categories of solutions: (1) efficient attention, (2) parameter-efficient image-to-video adaptation, (3) input token masking, and (4) multi-resolution patchification.

\vspace{-1.3em}

\paragraph{1. Attention mechanism.} Factorized attention~\citep[][]{arnab2021vivit, bertasius2021timesformer} separates the temporal and spatial dimensions over which self-attention is applied, reducing both memory and computational costs. However, this modification introduces a new temporal block within each transformer layer making initialization and model tuning more challenging. In contrast to~\cite{arnab2021vivit}, that initializes the new blocks with zeroes, we find that we achieve best performance when initializing the temporal blocks with the same self-attention weights of ViT. However, we add a gating mechanism which acts as a residual connection between the self-attention blocks: $h =  h + tanh(\alpha)
h_{temporal}$. Here, $\alpha$ is a trainable parameter initialized to zero, that helps maintain the capabilities of the original ViT during training.

\vspace{-1.3em}

\paragraph{2. Parameter-efficient adaptation.} We explore using parameter-efficient methods from NLP~\cite{chen2023peftdesignspace} to adapt image encoders to video, while only tuning a small percentage of model parameters. Most prior work adapts image-based models by freezing an image backbone and adding late, trainable temporal-fusion layers~\cite{yan2022video,cheng2023vindlu,zhang2023multimodal}. In contrast, we explore ways to use pre-trained image encoders and adapt them to \textit{video-first} architectures~\cite{pan2022st,liu2023revisiting,chen2022litevl}. Inspired by the success of parameter-efficient adaptation in NLP~\cite{zhang2023adaptivepeft}, we consider using MLP Adapters~\cite{houlsby2019parameter} and LoRA~\cite{hulora} (details in Appendix A). We also explore tuning only temporal self-attention blocks~\cite{chen2022litevl}, effectively as adapter layers, in factorized attention. In all variants, we still tune the video-specific 3D patch convolution.

\vspace{-1.3em}

\paragraph{3. Token masking.}

Most existing work samples videos at a fixed frames per second (FPS) rate~\citep[\eg,][]{arnab2021vivit,tongvideomae,yu2023self,alayrac2022flamingo}. However, semantics required for many video--language tasks vary slowly in the temporal dimension~\cite{zhang2012slow} and videos present high degree of redundancy between consecutive frames~\cite{tongvideomae}. We explore ways to sparsely sample the video input to reduce the number of input visual tokens. Specifically, we test random masking of input tubelet embeddings. Since consecutive frames are largely redundant, the same semantic signals could potentially be extracted even with high masking rates. For example,~\cite{tongvideomae} masks up to 95\% of the input video to reach optimal performance on the task of video-masked autoencoding. We demonstrate similar results in a video--language setting.


\vspace{-1.5em}

\paragraph{4. Multi-resolution patchification.} 

Finally, we test a simple approach to reduce redundancy in videos via more coarse-grained patchification in the temporal or spatial dimension, as commonly done in multiple-view video models~\cite{ma2022rethinking,feichtenhofer2019slowfast,yan2022multiview}. 
However, this decreases frame resolution, and may lose fine-grained information. As a result, we also experiment with TubeViT~\cite{piergiovanni2023rethinking} variant that combines flat patches and tubelets of different granularity to mitigate information loss. Following~\cite{piergiovanni2023rethinking}, we use four different convolution kernels that can encode either coarse-grained temporal or spatial information; details are in Appendix A.

\vspace{-0.5em}

\section{Datasets and Benchmarks} \label{sec:exp_setup}
\setlength{\tabcolsep}{5pt}

For contrastive pre-training, we use: (1) 27M video-text pairs (VTP) as described in \citep{alayrac2022flamingo}, (2) HowTo100M~\cite{miech2019howto100m} (HT100M; 100M instructional YouTube clips aligned with ASR using their timestamps, called HowTo100M Clips), and (3) VideoCC3M~\cite{nagrani2022learning} (3M video-text pairs based on Conceptual Captions~\cite{sharma2018conceptual}). 
Unfortunately, we find the text--video alignment in VideoCC3M to be of poor quality; instead, we use a modified variant with generated pseudo-labeled captions of every video by PALI \citep{chen2023pali} (see Appendices B, C). To pre-train with longer videos, we use a long version of HowTo100M (referred to as HowTo100M Summary) consisting of (1) the full-length videos with an average duration of 6.5 minutes and (2) their textual summaries generated by automatically cleaning and summarizing the ASR transcripts using an LLM~\citep[]{hoffmann2022training}. We also include the image datasets of~\citep{alayrac2022flamingo}. For video-to-text tuning, we use the same mixture of datasets but exclude HowTo100M Clips, since the noisy video-text alignments hurt performance.

We report text-video retrieval and captioning results on \textit{short video benchmarks}, with average video length $\leq$30 seconds: MSR-VTT~\cite{xu2016msr}, YouCook2~\cite{zhou2018towards}, ActivityNet Captions~\cite{krishna2017dense}, and VATEX~\cite{wang2019vatex}. To evaluate performance on longer videos, we consider video summarization on full-length versions of YouCook2 and ActivityNet Captions, with a video duration of up to 5 minutes, and multiple-choice video question answering (QA) on EgoSchema~\cite{mangalam2023egoschema}.

\setlength{\tabcolsep}{3pt}

\begin{table}[t]
\footnotesize
\centering
\begin{tabular}{@{}L{9em}cccccccc@{}}
\toprule
 & \multicolumn{2}{c}{MSR-VTT} & \multicolumn{2}{c}{VATEX} & \multicolumn{2}{c}{YC2} & \multicolumn{2}{c}{AN} \\
 \cmidrule(l){2-3} \cmidrule(l){4-5} \cmidrule(l){6-7} \cmidrule(l){8-9} 
 & T2V & V2T & T2V & V2T & T2V & V2T & T2V & V2T \\ \midrule
Joint ST-ViViT & 39.6 & \textbf{38.1} & 23.8 & \textbf{26.3} & \textbf{12.3} & \textbf{13.6} & 6.7 & 6.4 \\
Factorized ST-ViViT & \textbf{40.2} & 36.9 & \textbf{25.3} & 25.4 & 11.6 & 12.7 & 6.6 & \textbf{7.4}  \\
Avg Frame-level & 39.3 & 34.8 & 24.8 & 25.0 & 9.1 & 7.9 & \textbf{6.8} & 7.1 \\
Att-pool Frame-level & 38.4 & 37.5 & 21.9 & 26.1 & 9.0 & 8.9 & 6.1 & 6.2 \\
\bottomrule
\end{tabular}
\vspace{-1em}
\caption{Text-video retrieval results (\% Recall@1) when considering different visual backbones.}
\vspace{-1em}
\label{tab:ablation_visual_backbones}
\end{table}

\section{Experimental Results}

In Section~\ref{sec:results_exploration}, we describe our results evaluating alternatives in memory-efficient video encoder design; options described in Section~\ref{sec:methods_design_space}. For this analysis, we use ViT-B/BERT-medium, with training details in Appendix B and ablations on experimental design in Appendix C.

In Section~\ref{sec:main_results}, we combine our most competitive design choices from \ref{sec:results_exploration} and test our models on short and long video understanding benchmarks. We scale our best model variants to ViT-L/BERT-base with a 400M (or 1B) language decoder. We test our short video models on text-video retrieval and video captioning, and our long video models on video summarization and QA on 256-frame videos.

In Section~\ref{sec:results_benchmarks}, we share our experience working across short and long video benchmarks~\cite{xu2016msr, wang2019vatex, mangalam2023egoschema, das2013thousandyoucook1, caba2015activitynet}, offering insights about which ones yield robust temporal signal.

\begin{figure*}[t]
    \tiny
    \centering
    
    \hspace{-1em}
    \begin{subfigure}[b]{0.48\textwidth}   
        \tiny
        \centering 
        \includegraphics[width=0.95\textwidth]{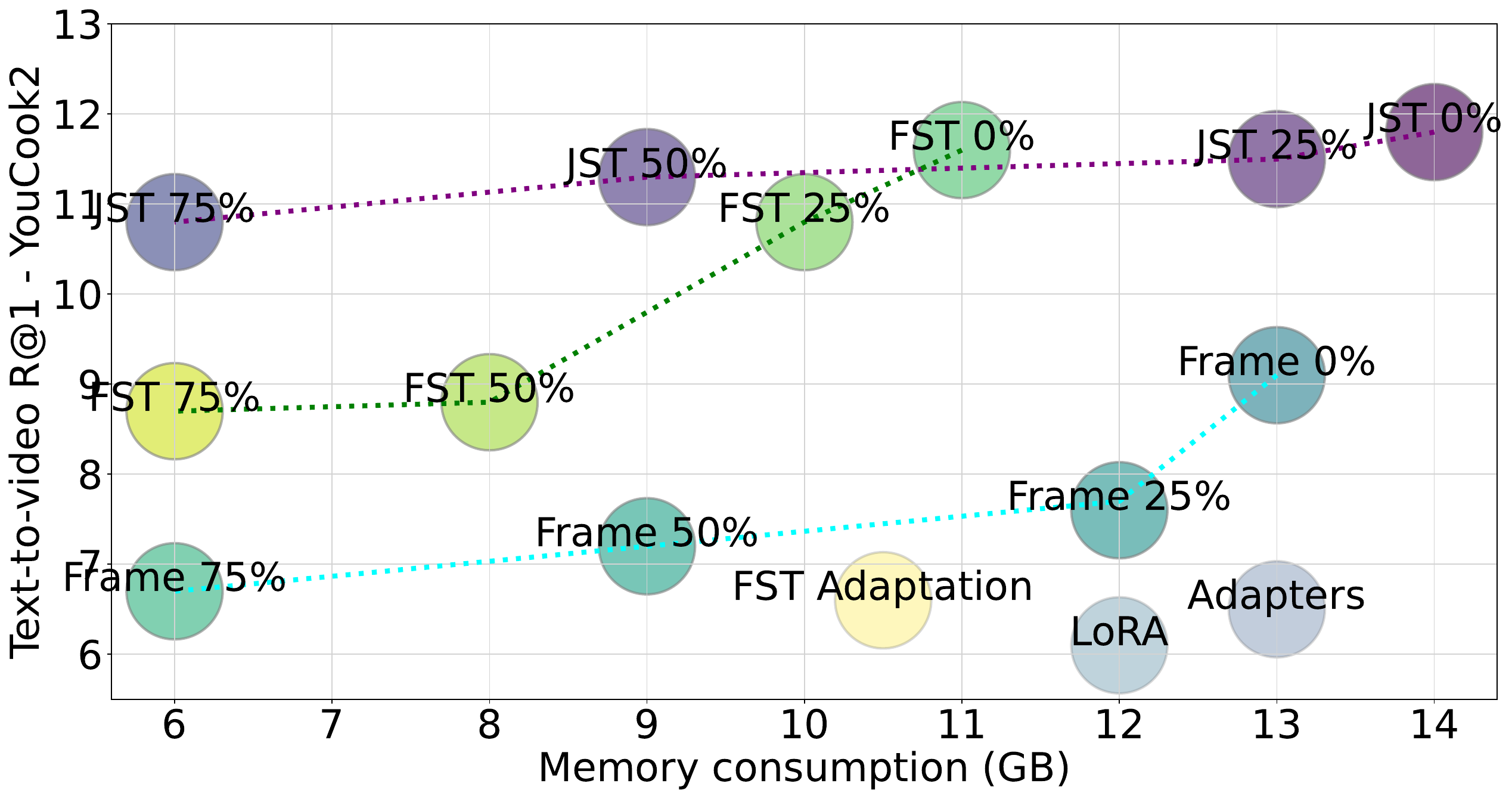}
    \end{subfigure}
    \hspace{2em}
    \begin{subfigure}[b]{0.48\textwidth}   
        \tiny
        \centering 
        \includegraphics[width=0.95\textwidth]{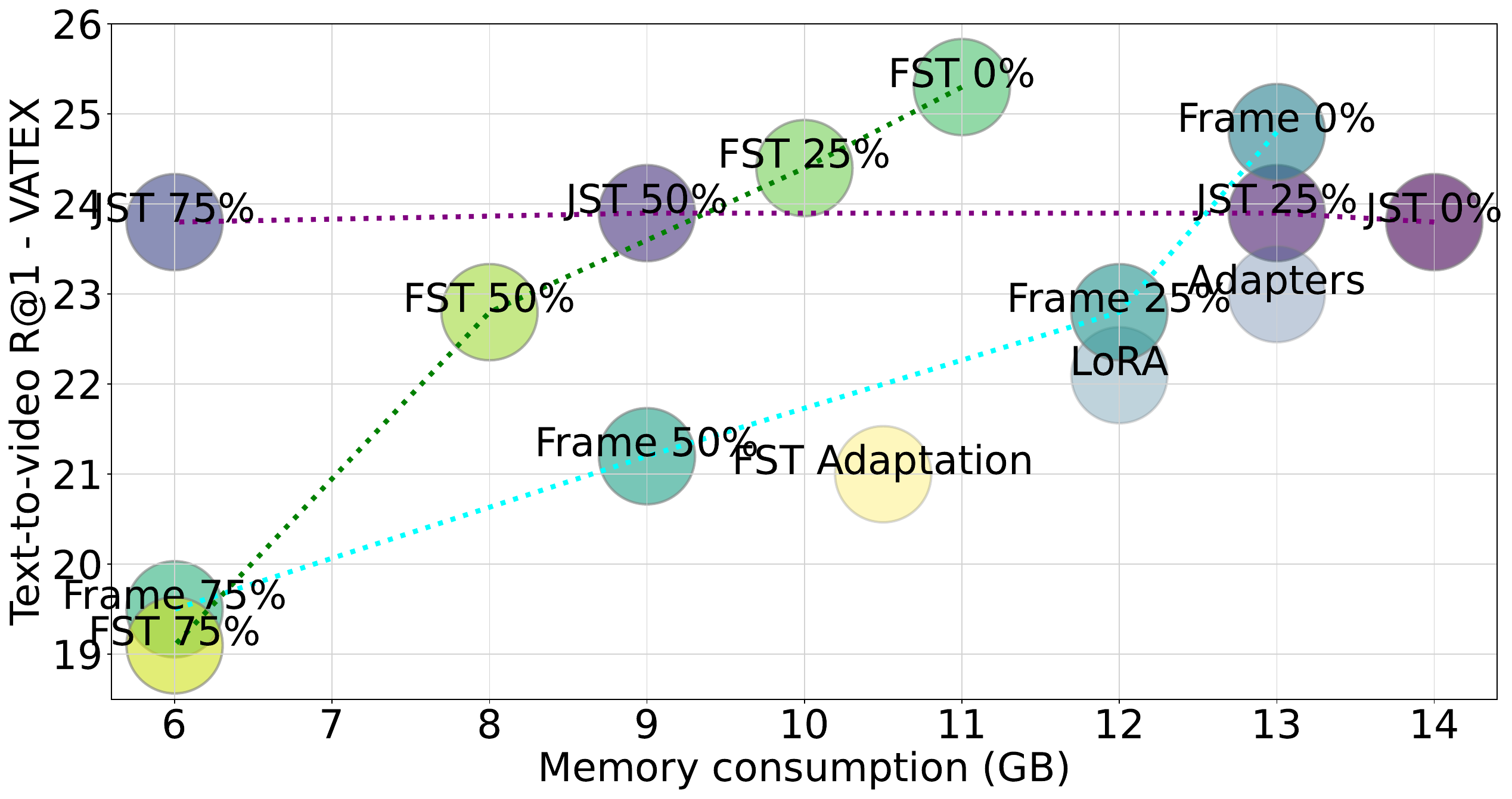}
    \end{subfigure}
    \caption[ ]
    {\small Trade-offs between performance (\% text-to-video Recall@1; y axis) and train-time memory consumption (x axis) for different backbones (joint space-time (JST), factorized space-time (FST), and drame-level encodings) with random input masking (0\% up to 75\%) or parameter-efficient methods for training (Adapters, LoRA, factorized temporal (FST) adaptation; lower opacity).}
    \label{fig:ablation_backbones_with_masking_vs_semifrozen}
\end{figure*}

\begin{figure}[t] 
        \tiny
        \centering
        \includegraphics[width=0.8\columnwidth]{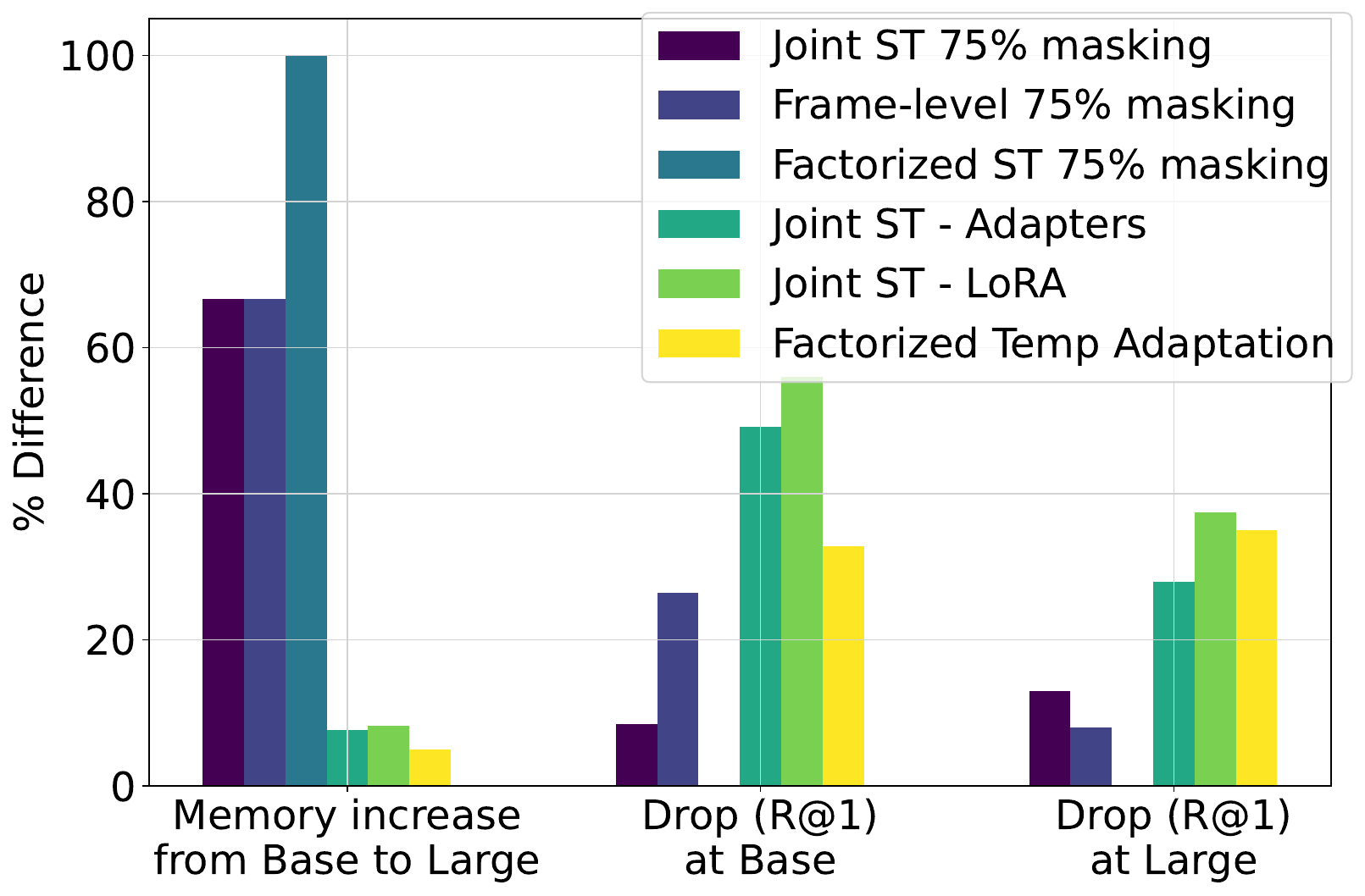}
\vspace{-1em}
        \caption[]%
        {{\small Difference (\%) in memory consumption for different model scales: (ViT-B vs ViT-L). We also report performance drop of efficient methods presented in Figure~\ref{fig:ablation_backbones_with_masking_vs_semifrozen} in comparison with the vanilla approach (i.e.,~no input masking and full fine-tuning) at different model scales to test whether behavior is similar.}} 
\vspace{-1.5em}
    \label{fig:ablation_backbones_with_masking_vs_semifrozen_large_model}
\end{figure}

\begin{figure*}[t]
    \tiny
    
    \centering
    \hspace{-3em}\begin{subfigure}[b]{0.31\textwidth}   
        \tiny
        \centering 
        \includegraphics[width=1.2\textwidth]{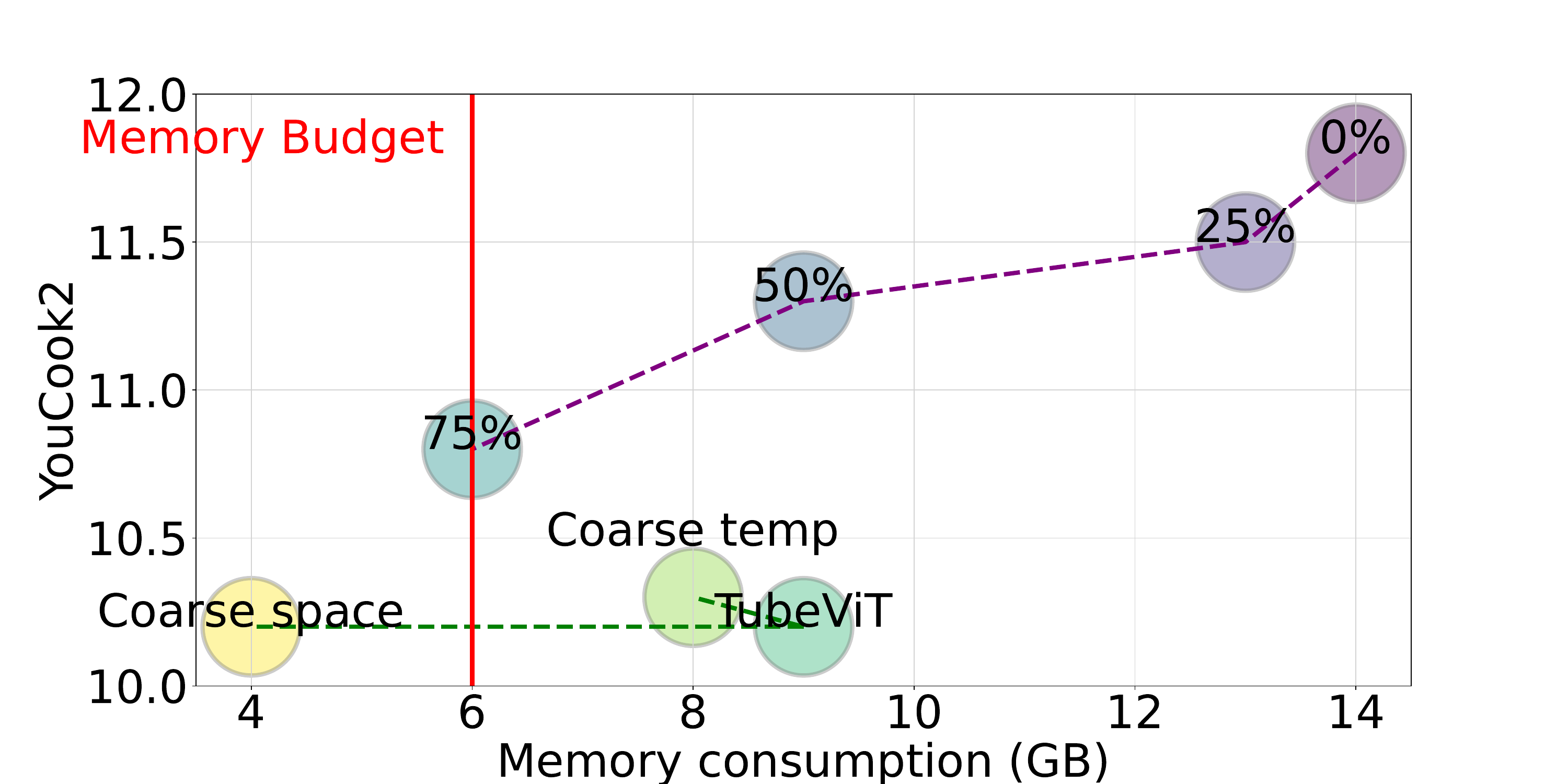}
    \end{subfigure}
    \hspace{3em}\begin{subfigure}[b]{0.31\textwidth}   
        \tiny
        \centering
        \includegraphics[width=1.2\textwidth]{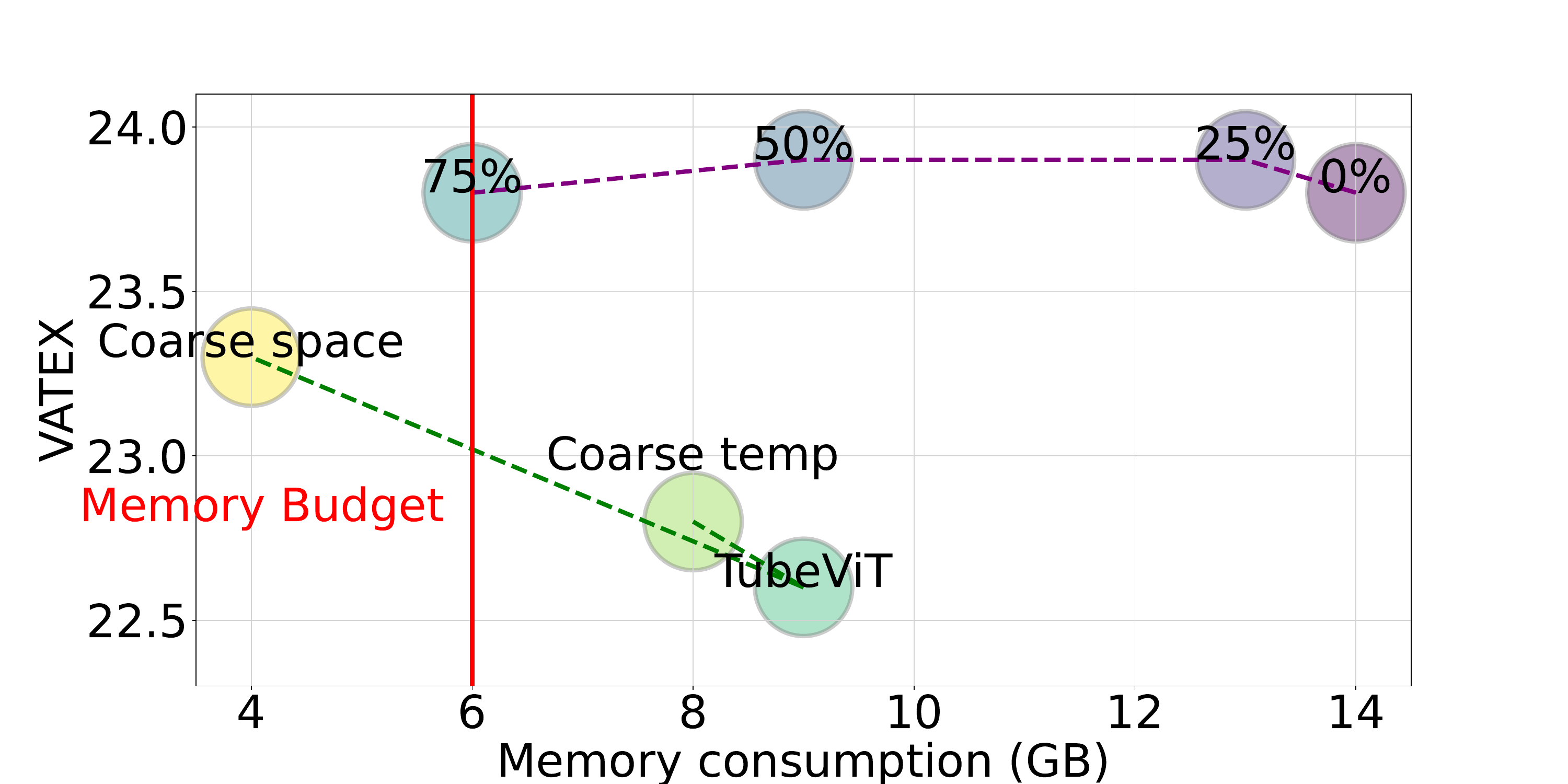}
    \end{subfigure}
    \centering
    \hspace{3em}\begin{subfigure}[b]{0.31\textwidth}   
        \tiny
        \centering
        \includegraphics[width=1.2\textwidth]{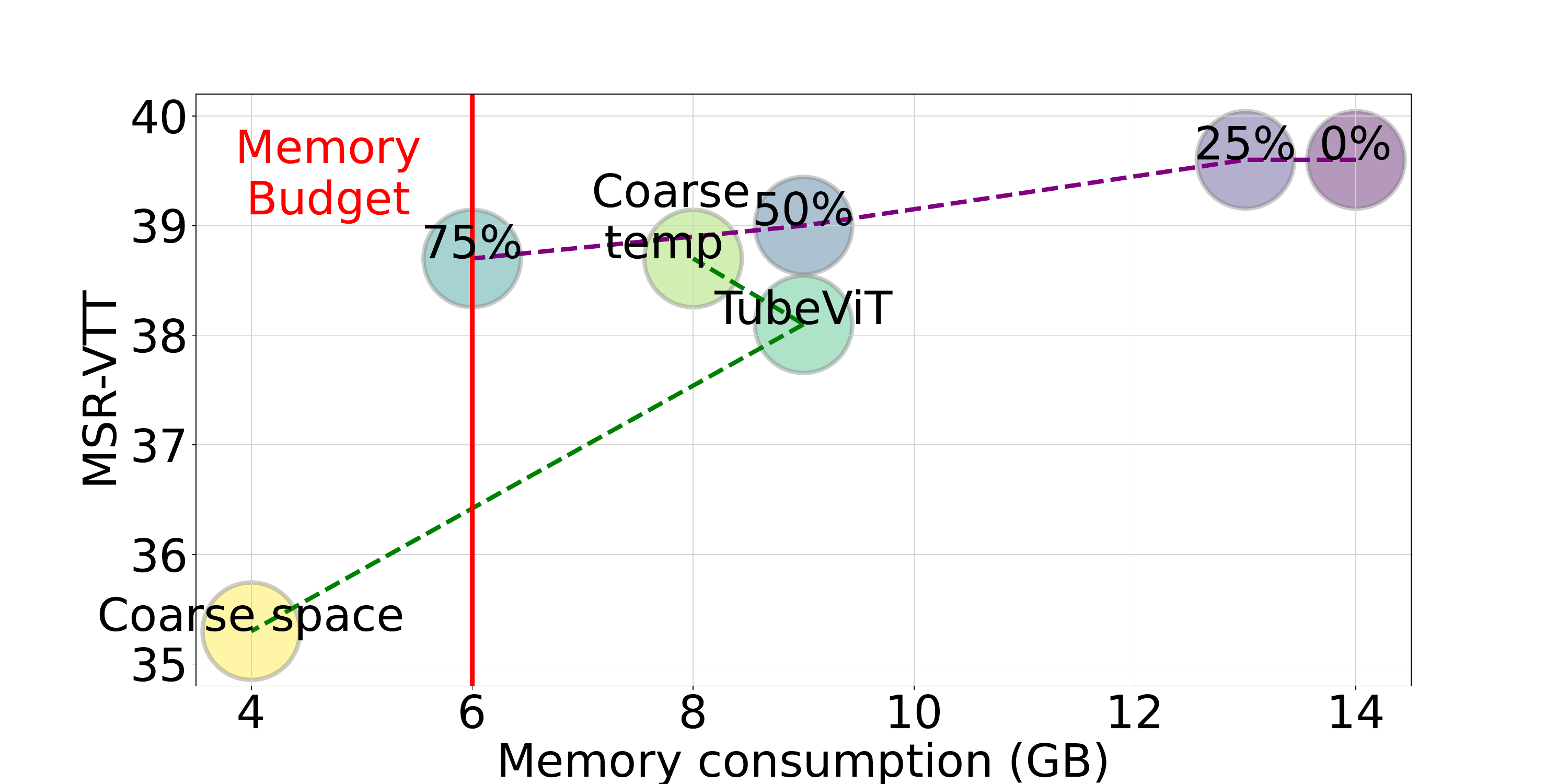}
    \end{subfigure}
\vspace{-1em}
    \caption[ ]
    {\small Trade-offs between performance (text-to-video Recall@1; y axis) and memory consumption (x axis) for input sampling methods: (1) high input masking ratios (0\% to 75\%) with joint space-time attention, (2) coarse-grained temporal (Coarse temp) and/or spatial (Coarse space) patchification with a fixed kernel and TubeViT which samples parts of the video with multiple 3D kernels of different granularity.}
\vspace{-1.5em}
    \label{fig:ablation_on_input_sampling_methods}
\end{figure*}

\begin{figure*}[t]
    \tiny
    \centering
    \includegraphics[width=0.9\textwidth]{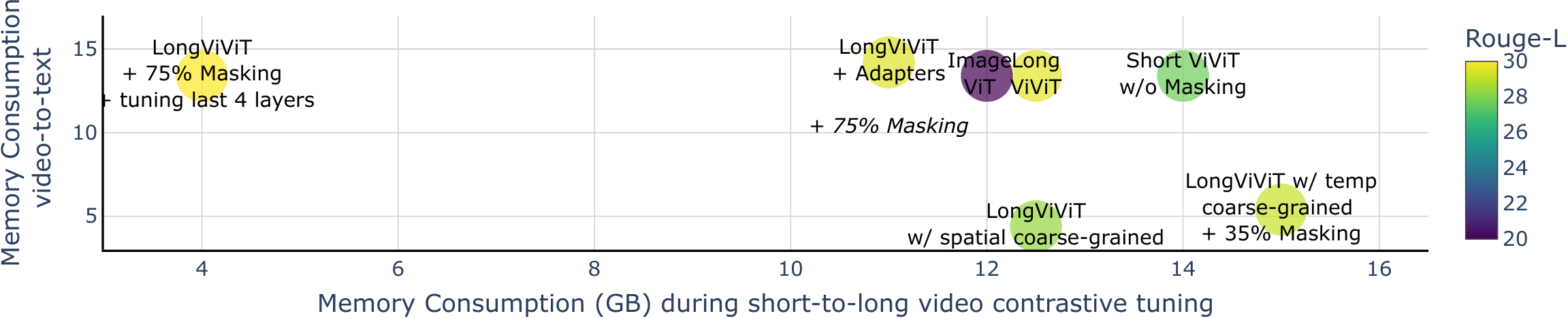}
    \caption[ ]
    {\small Scaling memory-efficient methods to more frames (i.e.,~128 frames) for ViViT-B and variants. We measure performance for video-to-text summarization on the full-length YouCook2 videos via Rouge-L (color-coded) while keeping track of memory consumption during short-to-long video contrastive tuning ($x$-axis) and video-to-text tuning ($y$-axis).}
    \label{fig:scaling_to_128_frames}
\end{figure*}

\subsection{Exploration of Memory-Efficient Designs} \label{sec:results_exploration}

We explore memory-efficient methods to train video-first encoders as described in Section~\ref{sec:methods_design_space}. We first consider short video inputs of 16 frames at 1 FPS and report peak train-time memory consumption vs. performance on text-video retrieval on short video benchmarks [\ref{sec:exp_setup}]. Then, we test whether our main findings hold for longer inputs (128+ frames) on video summarization on full-length YouCook2.

\vspace{-1.3em}

\paragraph{Base architectures.} 
We explore the memory/accuracy trade-off of different visual backbones in Table~\ref{tab:ablation_visual_backbones}: ViViT with joint space-time attention (\ie, Joint ST-ViViT), ViViT with factorized attention (\ie, Factorized ST-ViViT)~\cite{arnab2021vivit}, and frame-level (ViT-based) image encodings with average or attentional pooling (`att-pool')~\cite[][]{alayrac2022flamingo,yan2022video}.
Different methods perform similarly, especially on MSR-VTT and ActivityNet (AN). Interestingly, attentional pooling on top of frame-level encodings does not improve performance. ViViT with either joint or factorized attention performs best and presents higher gains for YouCook2 (YC2), 
the more temporally challenging benchmark [\ref{sec:results_benchmarks}]. In contrast to prior work~\citep[\eg,][]{yan2022video,cheng2023vindlu} which tests frozen image-to-video transfer and claims joint attention to be inferior, we find it to be competitive in this fully fine-tuned setting.

\vspace{-1.3em}


\paragraph{Architectures and token masking.}  We now test robustness of backbones when masking part of the input tubelets (0-75\%). We report Recall@1 on text-to-video retrieval for YouCook2 and VATEX\footnote{We do not observe significant sensitivity to input masking for MSR-VTT and ActivityNet Captions across all configurations (Section~\ref{sec:results_benchmarks}).} per backbone for different masking ratios in Figure~\ref{fig:ablation_backbones_with_masking_vs_semifrozen}. Joint space-time attention (JST) is robust against noise from masking up to 75\% during pre-training. The same \textit{does not} hold for frame-level encodings and factorized attention (FST), where performance drops consistently as we increase masking. We conclude that JST can better handle noisy inputs and use it in further exploration.

\vspace{-1.3em}

\paragraph{Parameter-efficient adaptation.} We next report performance of parameter-efficient image-to-video adaptation in Figure~\ref{fig:ablation_backbones_with_masking_vs_semifrozen}. We consider (1) JST with (a) MLP \textit{Adapters} at every layer of the encoder, (b) \textit{LoRA} with rank decomposition matrices in the self-attention and feed-forward transformer blocks, and (2) \textit{factorized temporal adaptation} where we tune the temporal self-attention. 
No adaptation method can reach the memory savings provided by high input masking, since we tune parameters \textit{depthwise} and gradient computation still requires backpropagation through the model. At the same time, we see significant performance drop, suggesting that adaptation of spatial-only models to the temporal dimension cannot be sufficiently addressed in semi-frozen fashion. Comparing parameter-efficient methods, we find MLP Adapters to be more competitive than LoRA, which is now canonical for LLMs. We hypothesize that LoRA is successful for tuning very small portions of the network and performing ``easier'' in-modality transfer.

\vspace{-1.3em}

\paragraph{Adaptation at scale.} We next scale from ViT-B/86M to ViT-L/307M in Figure~\ref{fig:ablation_backbones_with_masking_vs_semifrozen_large_model} and test whether observations hold with different model scales. We present the \% memory increase from base to large (left bar set) and \% performance \textit{decrease} of each method at each scale\footnote{Performance drop for factorized ST is omitted since the variant without masking leads to out of memory issues.}. Joint ST exhibits a similar memory pattern to frame-level, while leading to smaller accuracy drops, whereas factorized ST presents significant memory overhead with model scale due to the extra temporal parameters. For this reason, we exclude factorized ST from further experimentation. Finally, parameter-efficient methods are unable to achieve competitive performance at both model scales, although their memory requirements scale better with model size.

\vspace{-1.3em}

\paragraph{Multi-resolution patchification.} Given the outsized memory impact of input token count in Figure~\ref{fig:ablation_on_input_sampling_methods}, we additionally analyze: (1) \textit{coarse-grained patchification} in the temporal (convolution over 4 instead of 2 frames) and/or spatial (convolution over 32x32 instead of 16x16 pixel spaces) dimension, and (2) the \textit{TubeViT}~\cite{piergiovanni2023rethinking} approach of multiple tube kernels of different spatiotemporal size and strides.  For all benchmarks, masking the input at high ratios while maintaining a fine granularity of tubelets decreases performance significantly less than other input processing methods. Temporal coarse-grained patchification negatively affects benchmarks with richer temporal dependencies (i.e.,~YouCook2, VATEX) more than spatial. The opposite trend holds for datasets depending on spatial understanding (i.e.,~MSR-VTT, ActivityNet Captions\footnote{Omitted from Figure~\ref{fig:ablation_on_input_sampling_methods} but follows same patterns as MSR-VTT.}). TubeViT acts as the middle ground between the two by employing multiple kernels, with some performance degradation across all benchmarks. However, it is not able to alleviate the negative effects caused by considering coarser-grained information and presents higher memory requirements due to the multiple convolutions. Overall, we find that high masking with Joint ST and small tubelets yields the strongest memory/performance curves.

\setlength{\tabcolsep}{4.9pt}

\begin{table*}[t]
\footnotesize
\centering
\begin{tabular}{@{}p{8em}cccccccccccc@{}}
\toprule
 & \multicolumn{3}{c}{MSR-VTT} & \multicolumn{3}{c}{VATEX} & \multicolumn{3}{c}{YouCook2} & \multicolumn{3}{c}{ActivityNet} \\
 & \multicolumn{2}{c}{Zero-shot} & \multicolumn{1}{c}{FT} & \multicolumn{2}{c}{Zero-shot} & \multicolumn{1}{c}{FT} & \multicolumn{2}{c}{Zero-shot} & \multicolumn{1}{c}{FT} & \multicolumn{2}{c}{Zero-shot} & \multicolumn{1}{c}{FT} \\
\cmidrule(l){2-3} \cmidrule(l){4-4} \cmidrule(l){5-6} \cmidrule(l){7-7} \cmidrule(l){8-9} \cmidrule(l){10-10} \cmidrule(l){11-12} \cmidrule(l){13-13}
 & T2V/V2T & C1/C2 & \multicolumn{1}{c}{C1} & T2V/V2T & C1/C2 & \multicolumn{1}{c}{C1} & T2V/V2T & C1/C2 & \multicolumn{1}{c}{C1} & T2V/V2T & C1/C2 & C1 \\
\midrule
 \imagevit-L & 30.9/\textbf{41.6} & 24.6/25.1 & 63.6 & 36.2/\textbf{42.9} & 37.9/39.4 & 61.1 & 18.2/16.8 & 14.5/16.5 & 95.9 & 20.6/18.2 & 16.3/17.7 & 41.1 \\
 \shortvivit-L & 31.9/38.9 & 32.7/32.9 & 63.1 & \textbf{37.8}/\textbf{42.8} & \textbf{43.6}/43.0 & \textbf{67.5} & \textbf{20.4}/\textbf{20.5} & \textbf{21.0}/22.1 & \textbf{131.9} & \textbf{21.3}/\textbf{18.9} & 25.2/26.1 & \textbf{44.8} \\
 Eff\shortvivit-L & 29.9/38.3 & \textbf{33.8}/\textbf{33.9} & \textbf{63.8} & 34.4/42.7 & 41.3/42.7 &  64.7 & \textbf{20.5}/\textbf{20.3} & \textbf{21.1}/21.7 &  127.1 & 20.1/17.7 & \textbf{27.0}/26.5 & 41.1 \\
 VideoCoCa-L~\cite{yan2022video} & \textbf{33.3}/-- & 24.3 & -- & -- & -- & -- & 18.9/--  & 20.7 & -- & 31.5*/-- & 17.4 & -- \\ \midrule
 VideoCoCa-2.1B & \underline{34.3}/\underline{64.7} & 27.1 & \underline{73.2}
 & \underline{53.2}/\underline{73.6} & 22.8 & \underline{77.8} &  20.3/-- & 34.3 & 128.0 & 34.5*/33.0* & 19.3 & 39.3 \\
 Flamingo-3B~\cite{alayrac2022flamingo} & -- & -- & -- & -- & 40.1 & -- & -- & \underline{55.8} & -- & -- & -- & -- \\ 
\bottomrule
\end{tabular}
\vspace{-1em}
\caption{We present three model variants: \imagevit-L, that uses frame-level encodings with a late temporal fusion trained on images and videos, \shortvivit-L, our best performing video-first model with joint space-time attention, and Efficient \shortvivit-L (Eff\shortvivit-L) where we apply 75\% train-time masking for 3x memory savings. We also report performance for SoTA image-first models: VideoCoCa-L and Flamingo-3B, although they are bigger and not directly comparable. We report Recall@1 for zero-shot text-to-video (T2V) and video-to-text  (V2T) retrieval, and CIDEr for zero-shot and fine-tuned (FT) captioning when considering a 400M (C1) or 1B (C2) frozen LM for generation. ActivityNet retrieval results marked with * are not directly comparable, as these models uniformly sample frames, whereas we use the first frames of the long video with a fixed FPS of 1 to match experimental settings across benchmarks.}
\vspace{-1em}
\label{tab:large_variant_zeroshot_results}
\end{table*}

\vspace{-1.3em}

\paragraph{Scaling to longer videos.} We now test the best methods from Figure~\ref{fig:ablation_on_input_sampling_methods} on 128 input frames (32.7k visual tokens). We select methods that are within a memory budget (\textcolor{red}{red} vertical lines) and would fit on a 16GB device when expanded to long videos (128+ frames). We contrastively fine-tune [\ref{sec:contrastive_pretraining}] our best performing video model (i.e., Joint ST referred to as \shortvivit) on sequences of 128 frames on HowTo100M Summary [\ref{sec:exp_setup}], as detailed in Appendix B. We refer to this model as \longvivit. Finally, we fine-tune \longvivit for text generation (Section~\ref{sec:video_to_text}) on the full-length YouCook2, and report Rouge-L in Figure~\ref{fig:scaling_to_128_frames}, measuring memory consumption during both long-context contrastive ($x$-axis) and video-to-text ($y$-axis) tuning.

Validating our previous results, \imagevit (frame-level encodings) trained on longer videos with 75\% masking\footnote{We start from \imagevit trained on short videos with no masking.} significantly under-performs  video-first models (10 R-L drop). \shortvivit without further HT100M Summary training performs better than \imagevit, but cannot match models adapted to longer videos. \longvivit improves performance by 1.8 Rouge-L points over \shortvivit. Comparing input masking with coarser-grained patchification\footnote{Using the same fine-grained \shortvivit model for initialization.} provides similar insights to the previous paragraph.

Finally, we test MLP Adapters~\cite{houlsby2019parameter} for tuning \shortvivit to longer videos and observe no performance drop in comparison with full fine-tuning. This provides further evidence that parameter-efficient methods can be used for ``easier transfers'' but not temporal adaptation of spatial-only models. One downside of MLP Adapters is that it increases parameter count during video-to-text tuning ($y$-axis in Figure~\ref{fig:scaling_to_128_frames}).
Thus, we also experiment with contrastively tuning only the last four layers of the model. With this, we observe a further 3x decrease in memory, since we tune the network \textit{widthwise} and excise early layer gradient computation. At the same time, there is no memory increase for video-to-text and no performance degradation. We conclude that this combination (high input masking and tuning the last layers) is an effective setting for longer video adaptation. Given the observed robustness to masking, to further decrease video-to-text memory, we also mask 30\% of the input video during training and inference without observing any drop in summarization performance (see Appendix C).

\subsection{Main Results} \label{sec:main_results}
\vspace{-0.3em}
\paragraph{Short video benchmarks.} We present our main results on short video benchmarks in Table~\ref{tab:large_variant_zeroshot_results}. We use ViT-L with BERT-base for contrastive pre-training (Section~\ref{sec:contrastive_pretraining}) and a 400M frozen LM for video-to-text tuning (Section~\ref{sec:video_to_text}). Our entire video-to-text model accounts for $\sim$900M parameters, although we additionally test scaling the frozen LM to 1B parameters ($\sim$1.5B total count). We report Recall@1 for zero-shot text-video retrieval and CIDEr for zero-shot and fine-tuned video captioning. We consider three model variants: frame-level encodings \textit{\imagevit}, \textit{\shortvivit}, and \shortvivit with 75\% masking that uses 2-3x less memory (referred to as \textit{Efficient \shortvivit}). We also report results for VideoCoCa~\cite{yan2022video} and Flamingo~\cite{alayrac2022flamingo}\footnote{Models are not directly comparable due to different pre-training datasets, model sizes, training regimes, and input resolution. For instance,~\cite{yan2022video} fully fine-tune the LM and report results for 576 × 576 frame resolution instead of 256 × 256.}.

Our results remain consistent with our earlier observations. Contextualizing only intra-frame dependencies coupled with late temporal fusion (\imagevit) leads to inferior performance for retrieval and captioning on benchmarks with richer temporal dependencies (YouCook2, VATEX) but performs better on retrieval on MSR-VTT which relies on spatial understanding. Video-first architectures further tuned on video datasets (substantially noisier than curated image ones) improve temporal capabilities at the expense of spatial. For Efficient \shortvivit, we find that masking 75\% of the input video causes a performance drop: an average of 1\% absolute difference on zero-shot retrieval and no significant difference on zero-shot captioning across all benchmarks. The efficient model still performs similarly or better than \imagevit, especially on captioning and temporally rich benchmarks (e.g., YouCook2, VATEX), while consuming significantly less memory. Finally, when scaling the frozen LM component from 400M to 1B (C1$\rightarrow$C2) for zero-shot video-to-text generation, we observe moderate improvements across benchmarks and variants.

We compare our results against large image-based models with SoTA performance on video benchmarks (second block of Table~\ref{tab:large_variant_zeroshot_results}). Although results are not directly comparable due to different experimental settings, we are competitive and achieve even better results for temporally rich benchmarks (i.e., YouCook2) on text-video retrieval for models of similar parameter count\footnote{Video-text retrieval results on ActivityNet Captions are not comparable since we are only considering the first 16 seconds of the video, whereas \cite{yan2022video} uniformly sample frames from the entire video ($\sim$180 seconds).}. Moreover, our models significantly outperform VideoCoCa on most video captioning benchmarks even when considering their much larger versions in the zero-shot setting. Finally, when fine-tuning our video-to-text models with the 400M LM, we are again able to match and surpass the performance of the larger VideoCoCa-2.1B in two out of four benchmarks.

\vspace{-1.3em}

\paragraph{Long video understanding.} 
We further tune \longvivit-L on 256-frame HT100M Summary videos and evaluate zero-shot/fine-tuned summarization (YouCook2, ActivityNet) and QA (EgoSchema released subset); this is shown in Table~\ref{tab:long_video_results}. We additionally report results of~\longvivit on Perception Test~\cite{puatruaucean2023perception} in Appendix D, where videos are short but can benefit from higher FPS.

We consider two families of models. 1. Models that take as input 256 frames (first block of Table~\ref{tab:long_video_results}): \imagevit and \shortvivit pre-trained on 16-frame clips, and \longvivit further trained on 256-frame clips. 2. Modular approaches from prior work (second block of Table~\ref{tab:long_video_results}): (a) SeViLA Localizer~\cite{yu2023self} for localizing important frames in the long video given a textual query which are then fed into \shortvivit for performing the task\footnote{We select 16 frames using the pre-trained localizer provided by~\citep{yu2023self}. For video summarization, we use synthetic summaries of the video generated by PALI+Bard as the textual query for retrieving frames.}, and (b) the popular paradigm of captioning video segments or frames and using an LLM to aggregate information and form coherent summaries or answer questions~\cite{zeng2022socratic,li2023videochat,lin2023mmvid}. 
We try the latter approach with \imagevit and \shortvivit, generating captions over 16-second video segments and then feeding the captions to the September 2023 release of Bard, a much larger LLM than the ones used in previous results. We caption clips using uniform video segmentation (every 16 seconds) or an oracle segmentation when available (i.e., we consider ground-truth start and end timestamps for different events within ActivityNet and YouCook2 videos). We also test substituting our small video models with PALI-3 (5B parameters) for frame captioning\footnote{We consider captions of key frames per 8 seconds of video.}. Finally, we reference the SoTA fine-tuned performance on ActivityNet and YouCook2, when using specialized models with pre-computed features by multiple networks, object detectors, and domain-specific vocabulary~\cite{yamazaki2023vltint}.

\begin{table}[t]
\footnotesize
\centering
\begin{tabular}{@{}L{10.5em}ccccc@{}}
\toprule
 & \multicolumn{3}{c}{Zero-shot} & \multicolumn{2}{c}{Fine-tuned}  \\
 \cmidrule(l){2-4} \cmidrule(l){5-6} 
 & AN & YC2 & ES & AN & YC2 \\ 
\midrule
 \multicolumn{6}{c}{Inference with 256 frames}  \\
\midrule
\imagevit & 14.4 & 4.6 & 40.8 & 23.8 & 29.4 \\ 
\shortvivit & 15.4 & 7.0 & 47.9 & \textbf{24.3} & 29.5 \\
\rowcolor{blue!10} 
\longvivit  & 15.2 & \textbf{20.3} & \textbf{56.8}  & 24.0 &  \textbf{30.6} \\ 
\midrule
 \multicolumn{6}{c}{Modular approaches with 16-frame video models}  \\ 
\midrule
SeViLA-to-\shortvivit  & 16.2 & 4.2 & 49.6  & \textbf{24.4} & 28.3  \\
\imagevit-to-Bard & 18.1 & 15.8 & 35.0 & 22.9 & 19.1 \\
\hspace{1em} + oracle segments  & 16.3 & 16.2 & -- & 22.7 & 22.1 \\
\shortvivit-to-Bard & 19.3 & 18.1  & 42.0 & 22.7 & 20.8 \\
\hspace{1em} + oracle segments  & 18.3 & 18.2  & -- & 22.7 & 24.7 \\ 
\midrule
PALI~\cite{chen2023pali} 5B-to-Bard & \textbf{22.0} & 19.9 & 44.8 & -- & -- \\ 
Blind Bard & -- & -- & 27.0 & -- & -- \\ 
SoTA~\cite{yamazaki2023vltint} & -- & --  & -- & \textit{36.9}  & \textit{34.6} \\
\bottomrule
\end{tabular}
\vspace{-1em}
\caption{Results on long video-to-text benchmarks. We report Rouge-L for zero-shot and fine-tuned video summarization on ActivityNet Captions (AN) and YouCook2 (YC2) and zero-shot accuracy (\%) for multiple choice QA on EgoSchema (ES).}
\vspace{-2em}
\label{tab:long_video_results}
\end{table}

Looking through Table~\ref{tab:long_video_results}, we find that on ActivityNet, which contains less temporal dependencies~[\ref{sec:results_benchmarks}], modular approaches via frame selection or LLM-based aggregation of information (second block) perform well. Frame captioning via PALI combined with the power of LLMs is enough for the task in a zero-shot setting. For fine-tuned models, feeding either the long input or selected frames into \shortvivit perform better than utilizing Bard. On ActivityNet, we see no benefit from training further on longer videos.

In contrast, we find that short video and modular models are insufficient for addressing video tasks with longer-range temporal dependencies (YouCook2, EgoSchema). Adapting \shortvivit to longer contexts (\longvivit) significantly improves performance and achieves the best scores across all comparison approaches. Using Bard as an information aggregator over individual clip captions cannot achieve competitive performance, even when considering an oracle video segmentation for YouCook2 (Lines 3 and 5 in the second block of~Table~\ref{tab:long_video_results}). 
Surprisingly, even using a much larger and more powerful image-based model (PALI) cannot reach \longvivit on YouCook2 and EgoSchema. Interestingly, selecting 16 key frames and feeding them into \shortvivit also outperforms Bard-based methods on EgoSchema and fine-tuned YouCook2.
This suggests there can be temporal dependencies in long videos that cannot be resolved even with an optimal event segmentation for the video, or be aggregated by LLMs given inprecise visual information. On such benchmarks, \longvivit demonstrates strong performance even without LLM assistance.

\subsection{Brief Notes on Video Evaluations}
\label{sec:results_benchmarks}
We briefly describe some of our findings on video evaluations. Firstly, we find that blind Bard is able to achieve SoTA results on the \emph{full set} of EgoSchema (no visual input; 33.9\% accuracy vs. 32.1\% for the best model in~\cite{mangalam2023egoschema}). Adding visual information from PALI into Bard increases performance to just 39.2\%. However, on EgoSchema's released \emph{subset}, performance of blind Bard is 27\%, which is much lower than PALI-to-Bard (44.8\%), suggesting that the subset contains questions that rely more on visual grounding than pure language reasoning, so we report numbers on the subset in Table~\ref{tab:long_video_results} and on the full set in Appendix~\ref{sec:appendix_d}. 

Figure~\ref{fig:ablation_benchmarks} details a simple ablation across other video benchmarks to quantify temporal richness. We test removing either video or image data from the training mix and measure the effect on performance (video-to-text Recall@1). We see a dramatic performance drop when removing video data for YouCook2 and VATEX (up to 75\%). ActivityNet and MSRVTT suffer more from the absence of image data, whereas non-video training influences performance in lesser degree (as little as 18\% for MSR-VTT).
%
We believe there's room for more fine-grained, temporal-focused video--language benchmarks in the community.

\begin{figure}[t] 
        \tiny
        \centering
        \includegraphics[width=0.85\columnwidth]{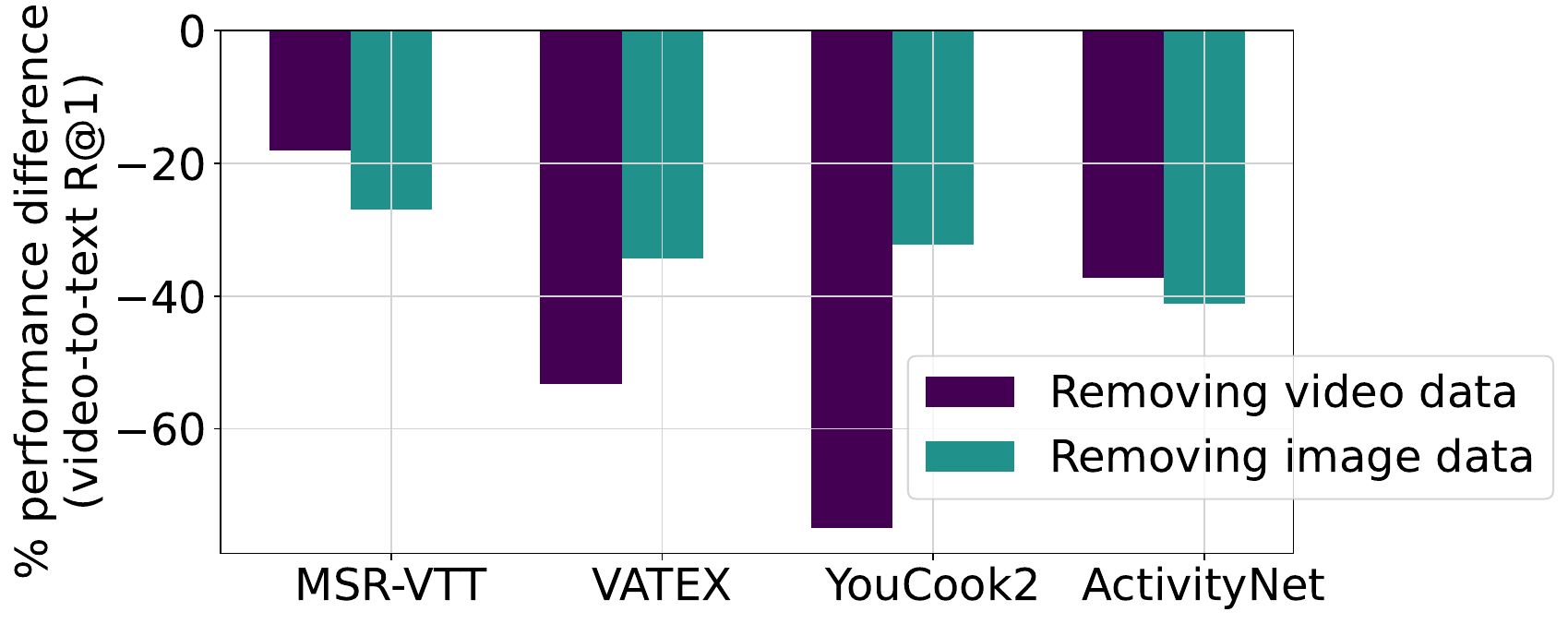}
\vspace{-1em}
        \caption[]%
        {{\small Performance difference (\%) per benchmark when we remove (1) video or (2) image data from the training mixture.}} 
\vspace{-3em}
    \label{fig:ablation_benchmarks}
\end{figure}
\section{Conclusions}

In short, we systematically analyze memory-efficient methods to scale video-first architectures to longer sequences of frames and demonstrate that just masking high percentages of the video ($\le$75\%) yields competitive results on long video--language tasks. Such masking shows a very small performance drop on short videos, provides 2-3x memory savings and allows scaling up to 4.3 minutes at 1 FPS (\longvivit) when freezing part of the short video network in our two-stage training.
\longvivit outperforms modular approaches with LLM assistance on video summarization and QA on benchmarks with richer temporal dependencies (YouCook2, EgoSchema). We overall demonstrate that encoding longer-range visual dependencies can make a difference in downstream performance and corrects mistakes that LLMs are unable to rectify.

\section*{Acknowledgements}
We thank Chuhan Zhang for her thoughtful feedback on early drafts of the work; Chris Dyer, Joao Carreira, and Gabriel Brostow for their support, feedback, and guidance through the project; and Lucia Lopez for her valuable help in generating synthetic video captions.
{
    \small
    \bibliographystyle{ieeenat_fullname}
    \bibliography{main}
}

\appendix

\appendix

\section{Methodology Details} \label{sec:appendix_a}

\paragraph{Image-to-video architectural adaptations.}
ViViT with joint space-time attention does not add any new parameters to vanilla ViT, which facilitates transfer between image and video models. However, we change the (position and patch) embedding layers to adapt to video inputs. In particular, we extend position embeddings to longer context via repetition, hence tokens within a frame have the pre-trained position encodings and tokens across time have identical position information. We experimented with both repetition and interpolation for initialization and find that both approaches provide similar results when fine-tuned on video data, whereas repetition works better in a zero-shot setting. As mentioned in Section 3.1 we use 3D convolution for embedding tubelets from the input video. We initialize the convolution weights via the 2D image-based weights by ``inflating'' them, \ie~replicating the filters along the temporal dimension and performing mean pooling~\cite{carreira2017quo,christoph2016spatiotemporal,arnab2021vivit}.

\paragraph{Multi-resolution Patchification and TubeViT.}
\citet{piergiovanni2023rethinking} tune multiple model variants of different sizes on different data mixtures (images vs. videos) in order to achieve good initialization of the multiresolution patch/tubelet embeddings. We aim to avoid additional pre-training steps with multiple pre-trained models and find that initializing all embedding layers with the image-based 2D weights works well in practice. Additionally, \citet{piergiovanni2023rethinking} handcraft fixed spatiotemporal position encodings to account for overlapping tubelets with different spatial and/or temporal strides, while showing that learnable position encodings lead to inferior performance. We overcome these obstacles and create a more generic approach that does not need handcrafting by employing factorized attention for processing the different ``views'' of the video (\ie overlapping parts of the video sampled at different spatiotemporal resolution). We find that this approach leads to better performance in contrast to flatten all multiresolution tubelets and feed them into the joint space-time attention as one long sequence.
Following~\cite{piergiovanni2023rethinking}, we use four convolution layers with the specified kernels, strides, and offsets. In our exploration, we also try spatiotemporal kernels of \texttt{(T,H,W)} sizes: (4, 16, 16,), (2, 32, 32), and (4, 32, 32).

\paragraph{Adapters and LoRA.} We explored use of MLP Adapters~\cite{houlsby2019parameter} and LoRA~\cite{hulora}.  For MLP Adapters, we add a bottleneck layer at every layer of the encoder (after the feed-forward block):
\begin{gather}
    h_{down,i} = f(LN(W_dh_i+b_d)) \\
    h_{up,i} = W_uh_{down,i} + b_u \\
    h_i = h_i + h_{up,i},
\end{gather}
where $W_d \in \mathbb{R}^{d_m \times d_b}$, $d_m$ is the model dimension, $d_b$ is the bottleneck dimension (\ie $\ll d_m$), $f(\cdot)$ is a non-linearity (we use $ReLU$) $LN$ is a trainable layer normalazation, $W_u \in \mathbb{R}^{d_b \times d_m}$, and $h_i$ is the output of the feed-forward layer.

For LoRA, we decompose the linear QKV input projection, the output self-attention projection, and the dense feed-forward block of each layer in ViViT: 
\begin{gather}
    h = W_o x + \frac{1}{\alpha}BAx,
\end{gather}
where $W_o$ is the original weight matrix of each block that remains frozen while tuning the learnable $B$ and $A$ matrices, $B \in \mathbb{R}^{d_m \times r}, A \in \mathbb{R}^{r \times d_m}$, $r \ll d_m$ is the rank of the decomposition matrices, and $\alpha$ is a hyperparameter for easier tuning of the model, as recommended by \cite{hulora}. 

For ViViT-B experiments: (1) for MLP Adapters, our adapter bottleneck dimension $d_b$ was set to 384 and we zero-initialize weights $W_d, W_u$ and biases $b_d, b_u$. (2) For LoRA, we use a $r$=64, and $\alpha = 1/64$ and same parameter initialization as \cite{hulora}. For ViViT-L experiments: the positioning and initialization of adapters were the same as with ViViT-B, but with $d_b$ of 768 (for MLP adapters) and $r$ of 128 (for LoRA).

\begin{table*}[t]
\footnotesize
\centering
\begin{tabular}{@{}L{10em}C{10em}C{10em}C{10em}C{10em}@{}}
\toprule
 & Image-to-Video Contrastive Pre-training & Short-to-long Video Contrastive Pre-training & Video-to-text Tuning  & Dataset-specific Fine-tuning \\ \midrule
Optimizer & \multicolumn{4}{c}{AdamW} \\
Learning rate schedule & \multicolumn{4}{c}{Cosine with linear warmup} \\
Gradient clip & \multicolumn{2}{c}{2.0} & \multicolumn{2}{c}{1.0} \\
Weight decay rate & \multicolumn{2}{c}{1e-2} & \multicolumn{2}{c}{1e-4} \\
Batch size & \multicolumn{2}{c}{512} & 128 & 64-128 \\
Base learning rate & \multicolumn{2}{c}{5e-5} & 4e-5 & 1-5e-6 \\
Linear warmup steps & \multicolumn{2}{c}{1k} & 2k & 1k \\
Training steps & 800k & 50k & 80k & 10k \\
Training steps & 800k & 50k & 80k & 10k \\
\bottomrule
\end{tabular}
\caption{Training specifications for (1) image-to-video contrastive pre-training, (2) short-to-long video contrastive tuning, (3) video-to-text tuning, and (4) dataset-specific fine-tuning for video captioning.}
\label{tab:training_details}
\end{table*}

\begin{table}[t]
\footnotesize
\centering
\begin{tabular}{@{}L{10em}ccc@{}}
\toprule
Training stage & Context & Compute & Duration \\ \midrule
Image-to-video contrastive pre-training & Short & 64 TPUv3 & 7 days \\
Short-to-long video contrastive pre-training & Long & 256 TPUv3 & 1 day \\
Video-to-text tuning with 400M LM & Short & 16 TPUv3 & 15 hours \\
Video-to-text tuning with 1B LM & Short & 64 TPUv3 & 2 days \\
Video-to-text tuning with 400M LM & Long & 128 TPUv3 & 2.5 days  \\
\bottomrule
\end{tabular}
\caption{Compute resources and duration for training our video-first encoders and video-to-text models used to report results in Tables 2 and 3.}
\label{tab:compute_details}
\end{table}

\paragraph{Temporal pooling + Perceiver resampler.} For video-to-text tuning, we use a Perceiver resampler with 3 layers, 1024 model hidden dimension, 8 heads for the multi-head attention blocks and 4096 inner-layer dimension for the feed-forward blocks. 
The Perceiver resampler was originally introduced by \citet{alayrac2022flamingo} in order to produce a fixed number of visual tokens to be fed into the LM independently of the output length of the visual encoder (e.g., for multiple images or videos). However, we empirically find that when we scale to videos beyond 16-32 frames, the Perceiver resampler becomes unstable during training leading to uniform attention distribution over visual tokens, even with appropriate Q/K cross-attention normalization and other tricks. In order to avoid unstable training for long frame sequences, we first average pool visual tokens across the temporal dimension in order to have a fixed number of tokens independently of the video length and then apply Perceiver resampler utilizing the same number of latent queries as the number of input tokens
(i.e.,~256 in our case for frames of 256x256 spatial resolution and convolution kernel of 16x16 spatial dimensions). Using this combination of a Perceiver resampler and temporal pooling leads to the best performance, compared to alternatives, although performance gains are generally small in comparison to completely removing the Perceiver resampler. We leave to future work the exploration of better ways to project and feed visual tokens into frozen LMs that can scale well with sequence length.


\section{Implementation Details} ~\label{sec:implementation}

In this section, we discuss implementation details including training specifications, compute resources, synthetic data for training on short and long videos, and evaluation details per benchmark.

\paragraph{Training details.}
We consider input frames with a 256x256 spatial resolution and patchify videos with a default convolution kernel of 2x16x16. We center-crop all frames and do not consider additional data augmentation or regularization methods. We present our experimental settings categorized by stage of training on Table~\ref{tab:training_details}.
We report compute requirements for training model variants on Table~\ref{tab:compute_details}. For offline evaluation, on all tasks and datasets, we used four TPUv5 chips.


\begin{table*}[t]
\footnotesize
\centering
\begin{tabular}{@{}L{10em}|c|cc|cc|cc|cc@{}}
\toprule
 & Dataset Size & \multicolumn{2}{c|}{MSR-VTT} & \multicolumn{2}{c|}{VATEX} & \multicolumn{2}{c|}{YouCook2} & \multicolumn{2}{c}{ActivityNet} \\
 & & T2V & V2T & T2V & V2T & T2V & V2T & T2V & V2T \\ 
 \midrule
ALIGN~\cite{jia2021scaling} & 1B & 24.1 & 17.1 & 6.2 & 3.3  & 1.0 & 0.5 & 1.9 & 1.1 \\
JFT~\cite{zhai2022scaling} & 300M & 16.2 & 14.6 & 7.6 & 6.3 & 1.6 & 1.0 & 1.5 & \underline{3.2}  \\
LTIP~\cite{alayrac2022flamingo} & 324M & 27.4 & 21.0 & 10.3 & 5.3 & 2.7 & 1.3 & 3.4 & 2.4  \\
All Image & 1.6B  & \underline{31.1} & \underline{27.7} & \underline{18.6} & 10.9 & 3.7 & 2.4 & \underline{4.8} & \underline{3.2}  \\
VTP~\cite{alayrac2022flamingo} & 27M & 23.2 & 21.3 & 13.3 & 15.3 & 2.5 & 1.9 & 3.3 & 3.1 \\
HowTo100M Clips~\cite{miech2019howto100m} & 100M & 14.3 & 13.7 & 5.6 & 6.6 & \underline{9.6} & \underline{9.9} & 1.2 & 1.5 \\
VideoCC3M~\cite{nagrani2022learning} & 7M & 12.3 & 11.9 & 5.0 & 4.4 & 1.0 & 0.0 & 1.1 & 1.0 \\
All Video & 134M & 26.2 & 24.7 & 14.9 & \underline{15.3} & 7.7 & 6.5 & 4.3 & 3.0 \\
All & 1.8B & \underline{36.3} & \underline{33.8} & \underline{20.6} & \underline{23.3} & \underline{9.3} & \underline{9.6} & \underline{6.2} & \underline{5.1} \\ 
\bottomrule
\end{tabular}
\caption{Performance of video model per pre-training image and/or video dataset on zero-shot text-to-video (T2V) and video-to-text (V2T) retrieval (\% Recall@1) when trained from scratch. Two best variants are \underline{underlined} per benchmark.}
\label{tab:ablation_pretraining_datasets}
\end{table*}

\paragraph{Synthetic training data.}
As mentioned in Section 5, we find that VideoCC3M presents poor video--text alignment and compromises performance of our models. For this reason, we experiment with generating synthetic textual descriptions of the videos via PALI-3~\cite{chen2023pali}. Videos in VideoCC3M are at 10FPS and on average only 10 seconds long and static; differences between consecutive frames are small. We roughly sample the center frame from each video and feed it into PALI for generating a detailed description (i.e. selecting the 50th frame, and if this fails, selecting the 25th frame, or if this fails, the 0th frame). We empirically find that (1) generated captions are more accurate than the original, automatically mapped ones, and (2) PALI-3 is able to generate long and detailed captions that mention several details present in the video. We show the effect of adding the PALI-captioned version of VideoCC3M in Appendix~\ref{sec:first_ablations}.

Moreover, we use the full-length videos of HowTo100M for training \longvivit on longer contexts (HowTo100M Summary; Section 5). The full-length videos have an average duration of 6.5 minutes and are accompanied by ASR, \ie automatic closed captions of people describing their actions. However, instead of directly using ASR, which can be noisy, not coherent between utterances, and contains irrelevant information and comments from the speakers, we use a LLM, namely Chinchilla~\cite{hoffmann2022training}, for better cleaning and summarizing the ASR. We further filter generated summaries to discard repetitions. The resulting summaries are more coherent, condensed and describe the desired task and accompanied actions.

\paragraph{Evaluation details per benchmark.}

For our ablation studies on text-video retrieval (Section 6.1) we use the validation sets of all benchmarks.
We follow the settings of~\citet{yan2022video}, when applicable, for reporting our main results in Section 6.2. Specifically:
\begin{itemize}
    \item \textit{MSR-VTT}: We report results in Table 1 of the main paper on the full test set for text-video retrieval and on a subset of 1000 examples from the test set for video captioning. Each video is accompanied by 20 different captions, so we report average over the different captions.
    \item \textit{ActivityNet Captions}: We report results on the \texttt{val1} subset. For short video evaluation in Table 2 of the main paper, we consider the first 16 seconds (sampling frames at 1 FPS from the beginning of the video) of the 180-second videos for paragraph-video retrieval and segment-by-segment captions for video captioning. For long video evaluation in Table 3 of the main paper, we consider the raw, full-length videos of \texttt{val1} without ground-truth segmentation. There are videos with multiple captions, in which case we report the average.
    \item \textit{YouCook2}: We report results on the ground-truth segments and full-length videos of the validation set in Tables 2 and3 of the main paper, respectively. 
    \item \textit{VATEX}: We report results on the validation set for text-video retrieval and on the test set for video captioning. Each video corresponds to 10 different captions, so we report average scores.
    \item \textit{EgoSchema}: We report results on the released subset with annotations (500 examples in total). We report results for the full set in Section~\ref{sec:appendix_d} and discuss limitations. For reporting results on multiple-choice QA, a task that our model has not learned to perform, we follow \citep{mangalam2023egoschema} and InternVideo~\cite{wang2022internvideo} and train our video-to-text model on MSRVTT-QA for 5k steps for adapting to the task.
\end{itemize}

\section{Ablation Studies} \label{sec:first_ablations}

\subsection{Video--Language Pre-training}

Here, we provide more insights on video--language pre-training, including data mixtures, model initialization, input context lengths, and auxiliary losses. For our ablations, we use ViT-Base/BERT-medium.

\paragraph{Image and video data mixtures.} First, we present performance of our video model (i.e.,~joint space-time attention) when trained from scratch using different pre-training image and/or video datasets in Table~\ref{tab:ablation_pretraining_datasets}. We report Recall@1 for zero-shot text-video retrieval and assess the quality of different pre-training datasets based on downstream performance. Overall, the suitability of each dataset largely depends on the benchmarks. Some key observations in addition to our discussion in Section 6.3 can be summarized as follows:
\begin{enumerate}[label=(\alph*)]
    \item Dataset size alone was not a key factor for strong downstream performance (e.g.,~ALIGN with 1B examples (Line 1) vs. LTIP with 324M examples (Line 3) for image datasets, HowTo100M Clips with 100M examples (Line 6) vs. VTP with 27M examples (Line 5) for video).
    \item Domain match between train and inference time can be a catalyst for good performance, even in cases where vision--text alignment quality is poor (e.g.,~HowTo100M Clips for YouCook2 (Line 6, third column) in comparison with all other benchmarks). This becomes prominent in video datasets, which tend to be very domain-specific in comparison to image benchmarks.
    \item Training on videos is important for benchmarks with strong temporal dependencies, but not necessary for benchmarks that largely depend on spatial understanding. In particular, comparing LTIP and VTP which are of similar domain and quality (\ie collected in a similar way) but of different sizes (324M vs 27M, respectively), we find that MSR-VTT is benefited more by a larger image dataset, whereas video-level information is more crucial than dataset size for VATEX. This further supports our observations from Section 6.3. 
    \item Overall, image and video examples are complimentary for video understanding, across all benchmarks.
\end{enumerate}

\begin{figure}[t]
    \centering
    \includegraphics[width=0.9\columnwidth]{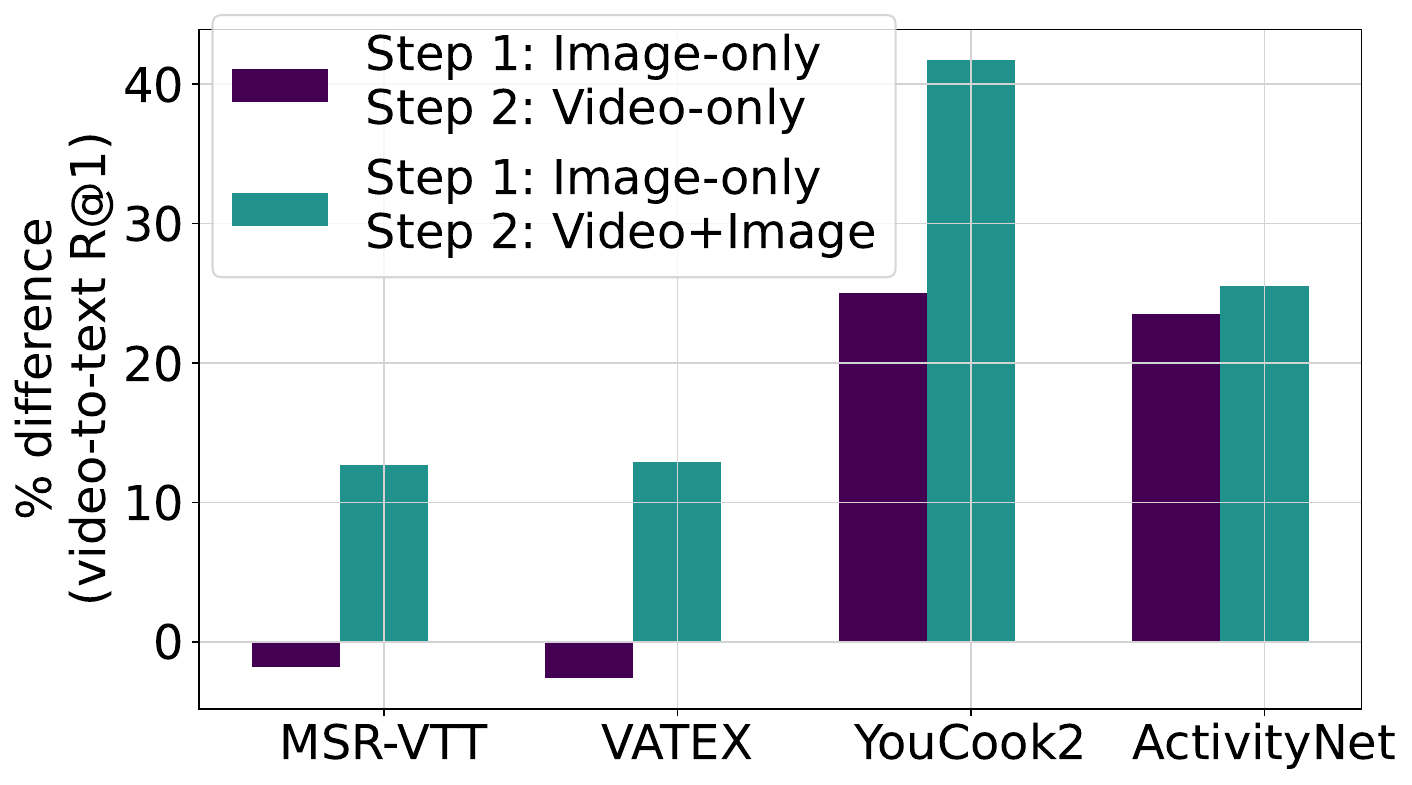}
    \caption{Performance difference (\% video-to-text Recall@1) of two step image-then-video training approaches in contrast to training video models on image and video data from scratch.}
    \label{fig:pretraining_image_video}
\end{figure}

\begin{table*}[t]
\footnotesize
\centering
\begin{tabular}{@{}L{18em}cc|cc|cc|cc@{}}
\toprule
 & \multicolumn{2}{c|}{MSR-VTT} & \multicolumn{2}{c|}{VATEX} & \multicolumn{2}{c|}{YouCook2} & \multicolumn{2}{c}{ActivityNet} \\
 & T2V & V2T & T2V & V2T & T2V & V2T & T2V & V2T \\ \midrule
Joint ST-ViViT (Contrastive only) & \textbf{39.9} & \textbf{38.1} & \textbf{23.9} & \textbf{26.3} & \textbf{11.4} & \textbf{12.6} & \textbf{6.7} & \textbf{6.4} \\
+ MLM loss & 39.0 & 36.6 & 23.0 & \textbf{26.4} & 11.0 & 12.0 & \textbf{6.8} & 6.0 \\
+ Captioning loss & \textbf{39.9} & 36.0 & 22.0 & 24.3 & \textbf{11.2} & 11.5 & \textbf{6.8} & 5.6  \\
\bottomrule
\end{tabular}
\caption{Text-video retrieval results (Recall@1) for different pre-training objectives. We apply 25\% masking of the video input in all cases.}
\label{tab:ablation_auxiliary_losses}
\end{table*}

\paragraph{Two step image-then-video training.} As discussed in Section 3.1, we first train an image ViT-based encoder on images, which we further tune on the video domain via joint space-time attention. This facilitates faster training and stronger contrastive objective due to larger batch sizes (i.e.,~8k vs 512). We test the effect of the two-step training approach on zero-shot retrieval in Figure~\ref{fig:pretraining_image_video}. We present \% difference on video-to-text Recall@1 when (1) training first on image-only and then video-only, or (2) training first on image-only and then image+video, where the same image datasets are used again with smaller weights for gradient computation, in contrast to jointly train on image+video from scratch. For case (1) we see improvements for two out of four benchmarks. While previous work only continue pretraining on video data~\cite{bertasius2021space,arnab2021vivit,yan2022multiview}, we also test keeping image datasets in the pre-training mix (case 2), and surprisingly, we observe an even larger relative improvement across all benchmarks. 

Given these observations we conclude that (a) there is indeed a benefit from pre-training first on images and then on videos, but (b) it is important to keep image samples in the mix, since video--text training is noisier and might have negative effect on spatial understanding, which can be mitigated in part from continual training on images. We confirm the latter by also directly evaluating on image benchmarks: for COCOCap~\cite{chen2015microsoft} an image-trained ViT achieves 35\% text-to-video Recall@1, whereas performance drops to 26\% for case (1) and 29\% for case (2).

\begin{figure}[t]
    \tiny
    \centering
        \includegraphics[width=0.9\columnwidth]{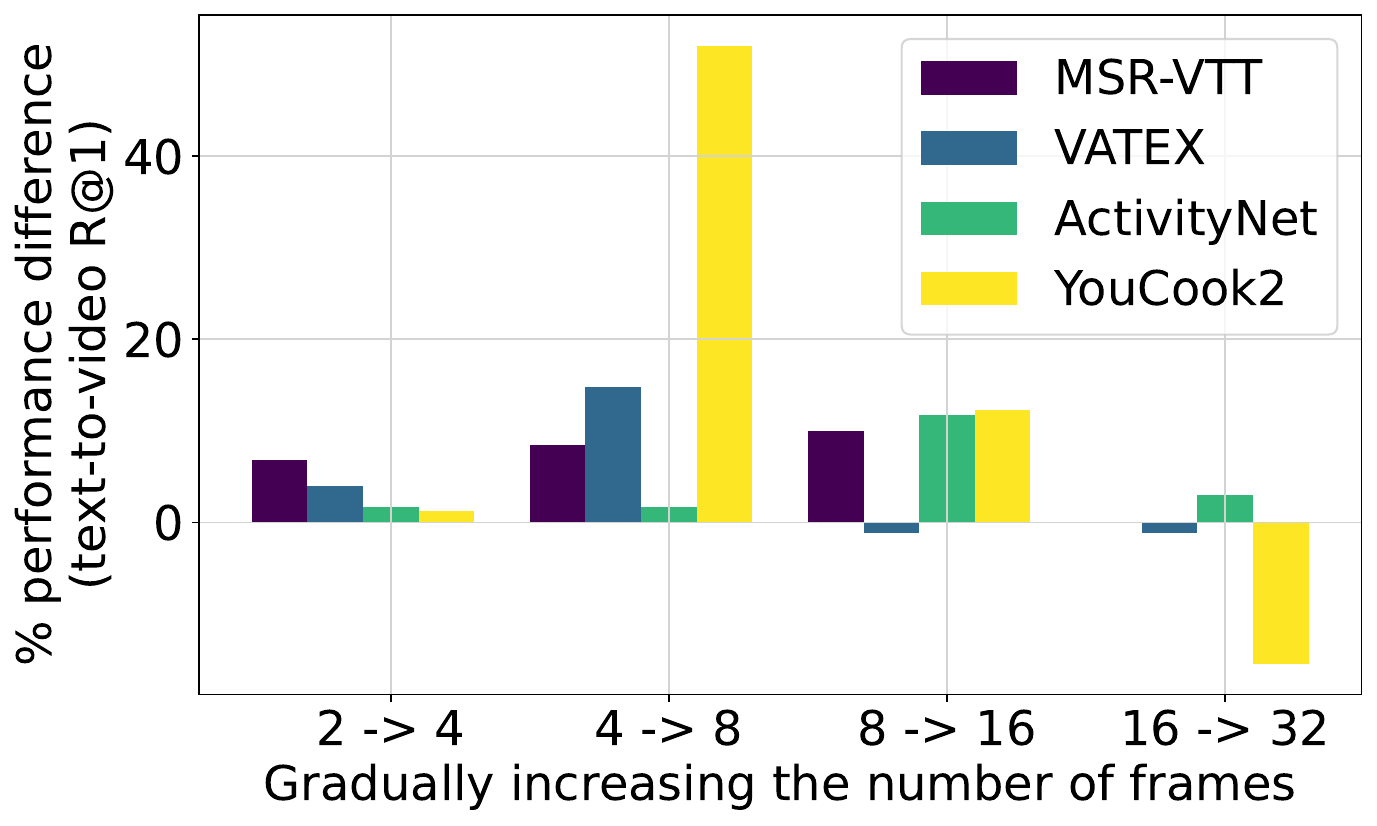}
    \caption[ ]
    {\small Performance difference (\%) for zero-shot text-to-video Recall@1 when gradually increasing the number of frames. This indicates the sensitivity of each benchmark to the number of frames and hence acts as an indicator of the temporal dependencies present in different benchmarks that are widely used. }
    \label{fig:num_frames_ablation}
\end{figure}


\paragraph{Context length.} 
We also experiment with variable number of frames per benchmark at a fixed FPS of 1. We range the number of frames from 2 up to 32 and present incremental performance difference as we gradually increase the number of frames in Figure~\ref{fig:num_frames_ablation}.
%
For most benchmarks, the performance improves as we increase the number of frames up to 16 frames, whereas longer inputs do not show benefits.  
%
%
%
This result highlights that most current benchmarks do not adequately measure temporal understanding (more than 16 frames). 
The most challenging dataset is YouCook2, where performance is close to random when considering less than 8 frames. VATEX also presents more challenging temporal dependencies with low performance when considering less than 4 frames.
These observations further validate our key findings of Section 6.3 and shed some light on which academic benchmarks are more appropriate to use for evaluating video models.

\paragraph{Auxiliary losses.} Although contrastive pre-training is a standard paradigm for image~\cite{radford2021learning,jia2021scaling} and video training~\cite{xu2021videoclip,miech2020end,luo2022clip4clip,alayrac2022flamingo}, prior work has explored captioning losses for training vision encoders in addition to or instead of contrastive objectives~\cite{tschannen2023image,li2022lavender,fu2021violet,zellers2021merlot,yan2022video}.
We next consider variants of the captioning loss for video pre-training as auxiliary losses to the contrastive objective with a 1:1 weighting between the two.

Adding a captioning loss in a dual encoder requires a multimodal encoder/decoder on top for fusing modalities and predicting tokens conditioned on the visual content. We add extra multimodal layers on top of the dual encoder similarly to~\cite{yan2022video}. We also consider two popular variants of the loss: (1) Masked Language Modeling (MLM), where we consider a multimodal \textit{encoder} and mask 15\% of the input textual tokens to predict, and (2) Captioning, where we consider a multimodal \textit{decoder} instead and predict each token of the caption autoregressively. We present results for the different pre-training objectives in Table~\ref{tab:ablation_auxiliary_losses}\footnote{We apply 25\% input masking for all variants for more efficient training.}. Overall, neither variant is able to improve results; in contrast performance mostly drops by adding the extra objective. Our results contradict prior work's observations on image pre-training~\cite{tschannen2023image} or video pre-training with frozen backbones~\cite{li2022lavender,fu2021violet,zellers2021merlot,yan2022video} and show that although contrastive objectives might be too coarse-grained for videos, considering captioning losses might be too fine-grained. 
We hypothesize that the very noisy video--text alignments hurt training of the video encoders when fully fine-tuned and the model needs to predict every textual token, which might not correspond to the visual input. In order to improve video pre-training in future work, we should look at either video-specific training objectives or better video--text alignments for video datasets.

\begin{figure}[t]
    \tiny
    \centering
        \includegraphics[width=\columnwidth]{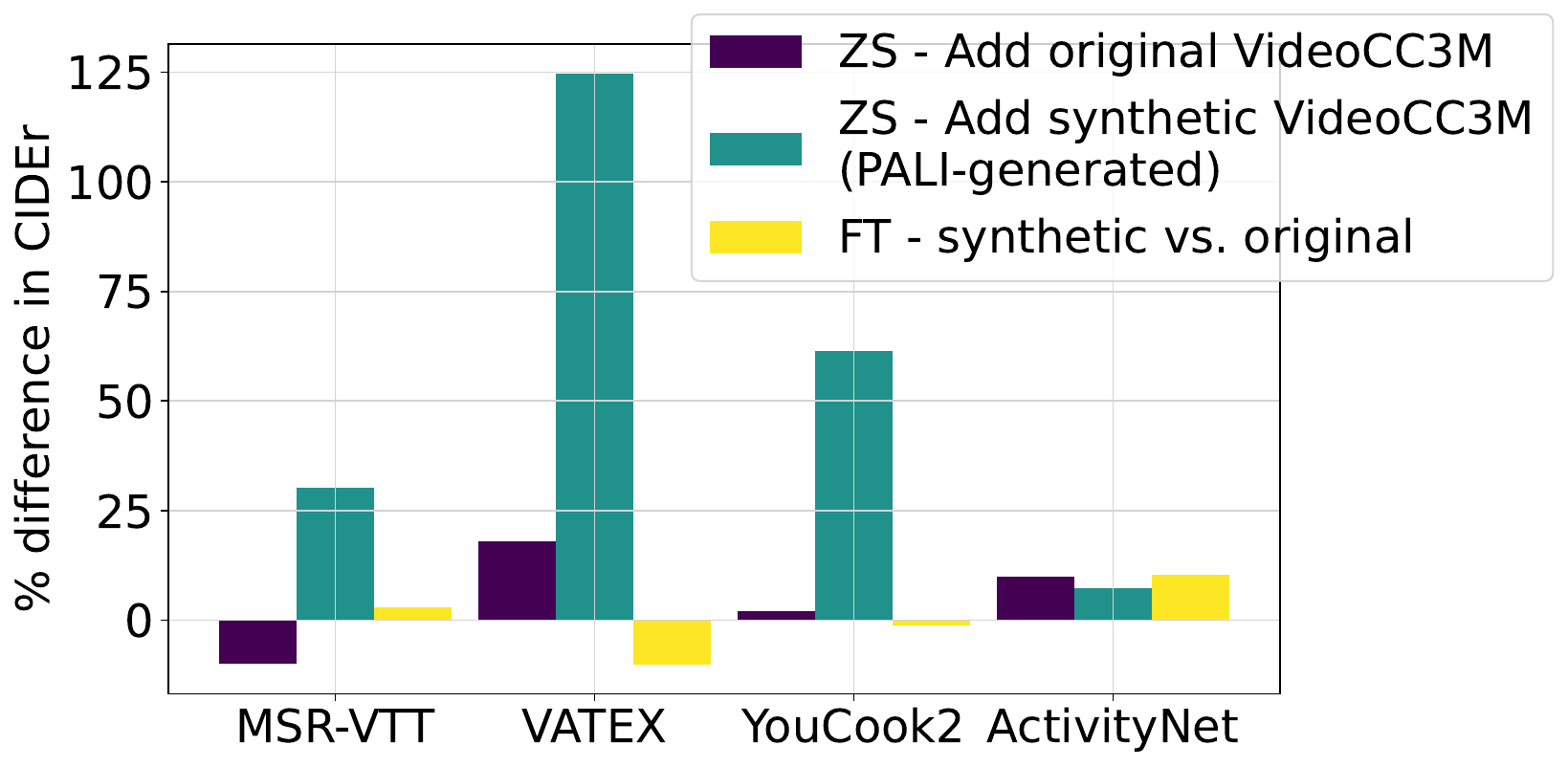}
    \caption[ ]
    {\small Performance difference (CIDEr) on zero-shot video captioning when we include either the original or synthetic VideoCC3M version (first two bars per benchmark). We also report performance difference between the two variants (i.e., synthetic vs. original dataset) when we fully fine-tune the models (third bar per benchmark).}
    \label{fig:synthetic_videocc3m}
\end{figure}

\subsection{Video-to-text Tuning}

\paragraph{Synthetic VideoCC3M}


We compare the original and synthetic versions of VideoCC3M (see Section~\ref{sec:implementation}) for video-to-text tuning in Figure~\ref{fig:synthetic_videocc3m}. Given a version of \shortvivit-L-to-text tuned only on image datasets from Table~\ref{tab:ablation_pretraining_datasets} and VTP for videos\footnote{We exclude HowTo100M Clips for short video-to-text tuning, since the dataset is very noisy and leads to drastic drops in performance across most benchmarks.}, we measure \% performance difference on CIDEr for zero-shot video captioning when we include either version of VideoCC3M (two first bars per benchmark). We observe a very large performance improvement (up to 125\% relative increase) when we use the synthetic version of VideoCC3M for three out of four benchmarks\footnote{Improvements for YouCook2 are smaller,~\ie less than 10\%.}. In contrast, using the original version of VideoCC3M provides moderate improvements for two out of four benchmarks (10-20\%), no improvement for YouCook2 and has a negative effect on MSR-VTT.

However, observations do not hold when we fine-tune different model versions.
In particular, we also report performance difference of the model trained with the synthetic dataset version against the one trained with the original one when we fine-tune them on the target datasets (third column per benchmark). In this case, we do not see benefit by using the synthetic dataset and even suffer performance drop for VATEX (10\% relative decrease). 

Overall, our findings for using synthetic video datasets are mixed. Our main hypothesis is that synthetic video captions can benefit model training in zero-shot settings since the model learns to produce \textit{longer} and \textit{more descriptive} captions (as empirically observed) which benefits metrics such as CIDEr. However, such benefits vanish when we further tune the model to the domain and style of interest.

\begin{figure}[t]
    \tiny
    \centering
        \includegraphics[width=\columnwidth]{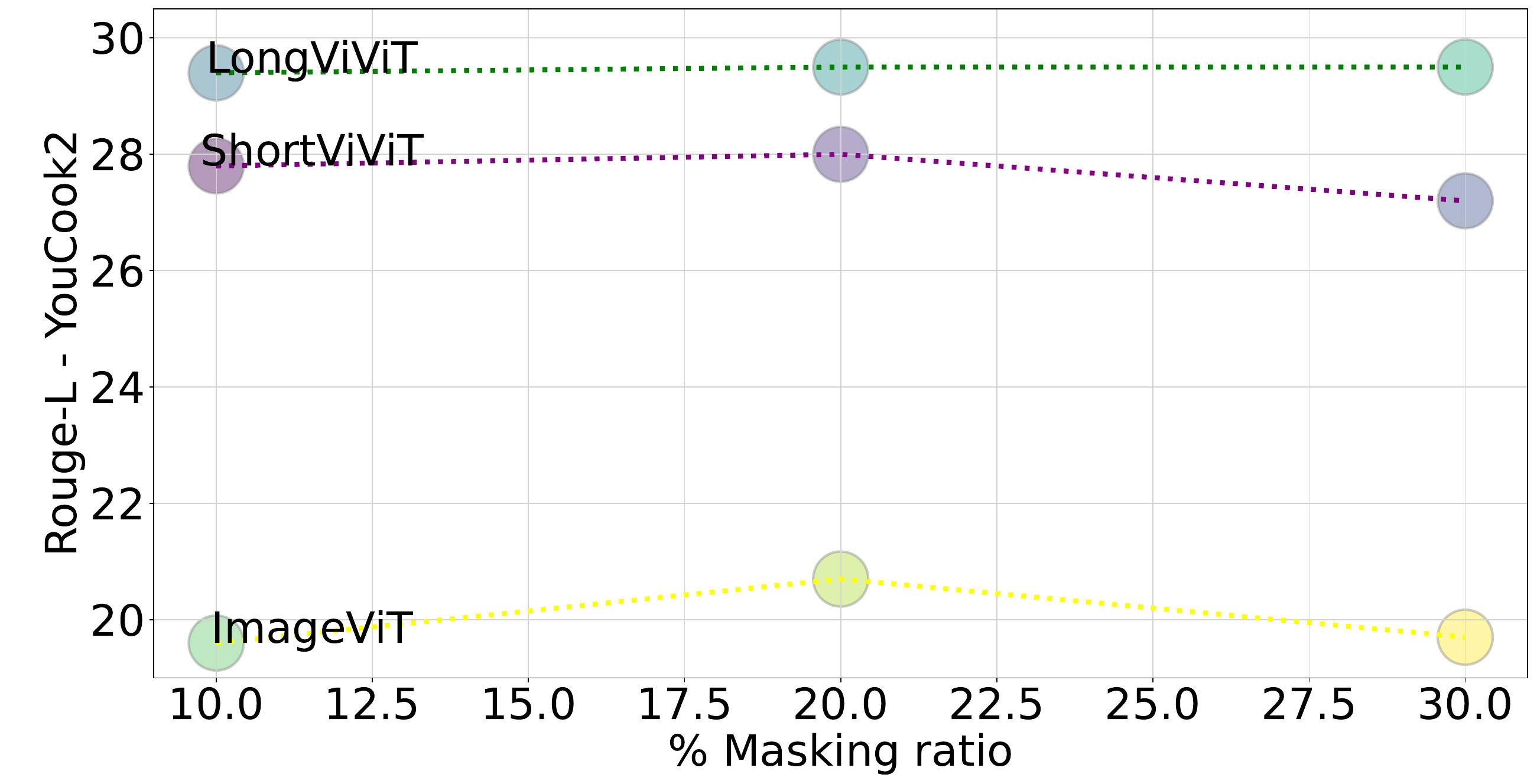}
    \caption[ ]
    {\small Rouge-L scores for long video models presented in Figure 5 of the main paper when applying different masking ratios during both training and inference ranging from 10 to 30\%.}
    \label{fig:masking_ratio_for_video_to_text}
\end{figure}

\paragraph{Video-to-text Masking}
We mention in Section 6.1 that we additionally apply up to 30\% masking for training and inference on video-to-text. In Figure~\ref{fig:masking_ratio_for_video_to_text}, we present performance (Rouge-L) on video summarization on YouCook2 for the setting described in Section 6.1 when we apply masking at different ratios (10-20-30\%) for three model variants: \imagevit, \shortvivit, and \longvivit (same models presented in Figure 5 of the main paper). Overall, we do not observe significant performance degradation when considering different masking ratios across all model variants, and hence we used 30\% masking for our main experimental results on longer videos (Section 6.2).

\section{Additional Experimental Results} \label{sec:appendix_d}

\begin{table}[t]
\footnotesize
\centering
\begin{tabular}{@{}L{10.5em}cc@{}}
\toprule
 & ES-Subset & ES-Full \\ 
\midrule
 \multicolumn{3}{c}{Inference with 256 frames}  \\
\midrule
\imagevit 1B & 40.8 & 30.9 \\ 
\shortvivit 1B & 47.9 & 31.0 \\
\rowcolor{blue!10} 
\longvivit 1B & \textbf{56.8} & 33.3 \\ 
\midrule
 \multicolumn{3}{c}{Modular approaches with 16-frame video models}  \\ 
\midrule
SeViLA-to-\shortvivit  & 49.6 & 31.3  \\
\imagevit-to-Bard & 35.0 & 35.0 \\
\shortvivit-to-Bard & 42.0 & 36.2 \\
\midrule
SeViLA~\cite{yu2023self} 4B & 25.7 & 22.7 \\ 
PALI~\cite{chen2023pali} 5B-to-Bard  & 44.8 & \textbf{39.2} \\ 
Blind Bard & 27.0 & 33.2 \\ 
Previous SoTA~\cite{wang2022internvideo,mangalam2023egoschema} & -- & 32.1 \\
\bottomrule
\end{tabular}
\caption{EgoSchema results (\% Accuracy on multiple-choice QA) for subset and full evaluation set.}
\label{tab:long_video_results_egoschema}
\end{table}

\paragraph{EgoSchema full evaluation set.}

We present our results on both the subset and full set of EgoSchema in Table~\ref{tab:long_video_results_egoschema}. Overall, we find that blind LLMs can answer a large percentage of questions without requiring any visual grounding (33.2\% for the full set in contrast to 27.0\% for the subset). Moreover, incorporating visual context to Bard via PALI captioning boosts performance only by 18\% for the full set in comparison to 66\% relative improvement for the full set. 
Hence, we find that models utilizing LLMs have an advantage for answering the full set questions and can achieve more competitive performance independently of the quality of the visual encodings. 
Finally, \longvivit still achieves the best performance for the full set when compared with models employing LMs of equal size, and the original SeViLA~\cite{yu2023self} model\footnote{In the original setting, SeViLA Localizer is trained with 32 input frames uniformly sampled from the video and selects 4 frames to feed to the Answerer.} fails to address the task despite its size (\ie 4B parameters).
Given these observations, we overall find the subset to be more challenging than the full set for video understanding evaluation and invite future work to also report performance on the subset for complete comparisons.

\paragraph{Perception Test.} We present results on  multiple choice video-QA task of the recently released benchmark, Perception Test~\cite{puatruaucean2023perception}. This benchmark is slightly out of the long video domain -- videos are relatively short ($<$30 seconds), and run at 10 FPS. Nonetheless, we hypothesize that given the nature of questions involving actions and localization, models could benefit from higher FPS and subsequently, longer sequences of input frames. We present results for our model variants and modular methods in Table~\ref{tab:long_video_results_pt}. Indeed, we are able to boost performance when processing videos at 5 FPS considering 256 input frames with \longvivit (\ie comparison of models in first block of Table~\ref{tab:long_video_results_pt}). Our model still performs better than modular methods that utilize Bard, including PALI-3 for frame captioning. We moreover outperform reported zero-shot results by Flamingo-3B. Finally, we are very close (0.5\% absolute difference) to SoTA SeViLA~\cite{yu2023self}, a model with 4B parameters and separately tuned localizer, which however fails to generalize to longer videos and questions about longer-range dependencies (\ie EgoSchema; Table~\ref{tab:long_video_results_egoschema}).

\begin{table}[t]
\footnotesize
\centering
\begin{tabular}{@{}L{20em}c@{}}
\toprule
  & PT \\ 
\midrule
\imagevit 1B (32 frames @ 1FPS) & 39.1 \\
\shortvivit 1B (32 frames @ 1FPS) & 41.9 \\
\rowcolor{blue!10} 
\longvivit 1B (256 frames @ 5FPS) & \textbf{45.7} \\ 
\midrule
\imagevit-to-Bard & 37.8 \\
\shortvivit 1B-to-Bard & 38.8 \\
\midrule
Flamingo~\cite{alayrac2022flamingo} 3B & 43.6  \\ 
SeViLA~\cite{yu2023self} 4B & \textbf{46.2} \\ 
PALI~\cite{chen2023pali} 5B-to-Bard & 42.4 \\ 
Blind Bard & 36.8 \\ 
\bottomrule
\end{tabular}
\caption{Accuracy (\%) on multiple-choice QA on Perception Test (PT). Models in the second and third blocks process videos at 5 FPS, except for Flamingo and SeViLA which follow the settings reported on~\cite{puatruaucean2023perception}.}
\label{tab:long_video_results_pt}
\end{table}

\begin{figure*}[t]
    \tiny
    \centering
    \hspace{-3em}\begin{subfigure}[b]{0.3\textwidth}   
        \tiny
        \centering 
        \includegraphics[width=0.85\textwidth]{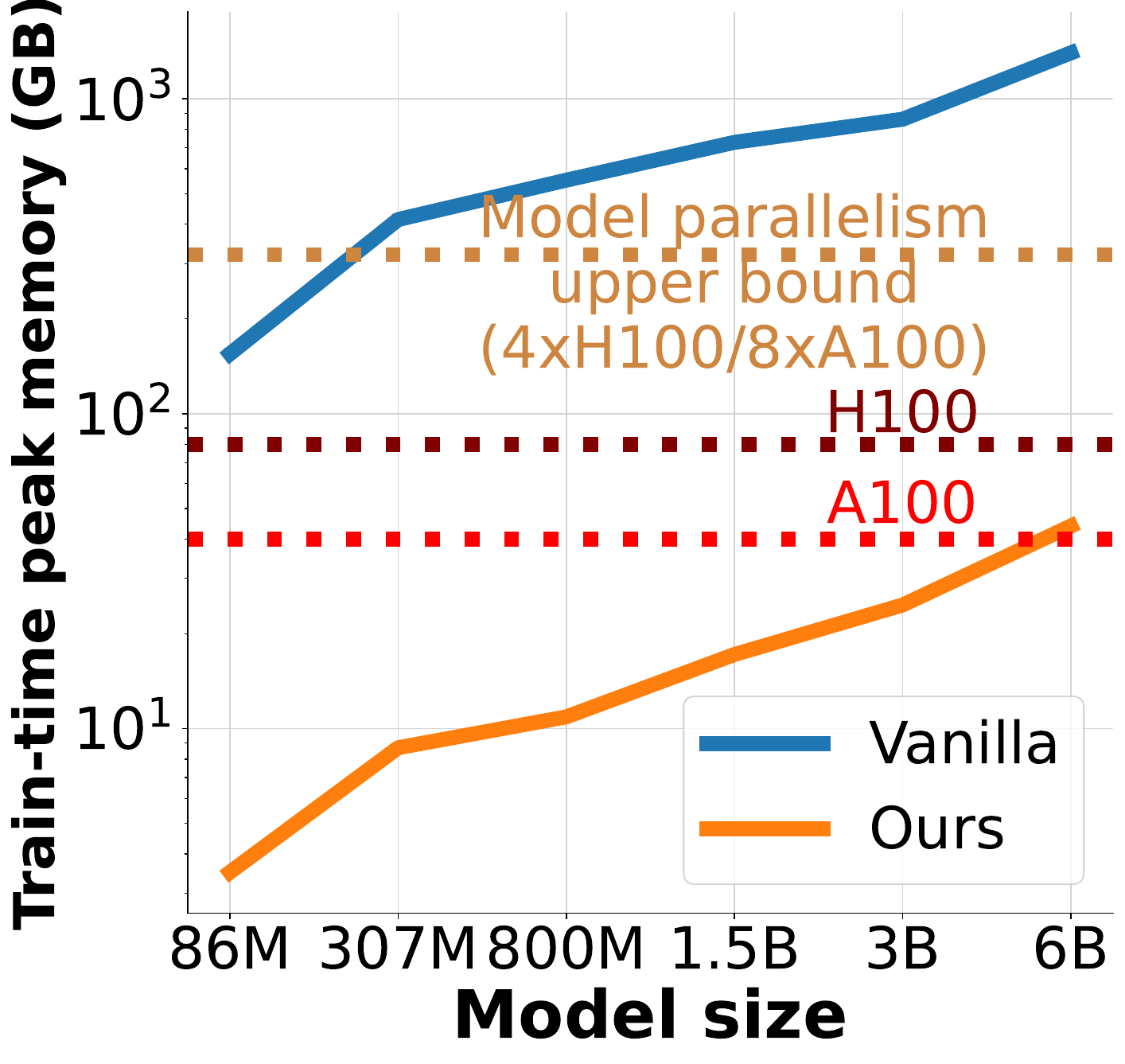}
        \label{fig:train_time_model_size}
    \end{subfigure}
    \hspace{3em}\begin{subfigure}[b]{0.3\textwidth}   
        \tiny
        \centering
        \includegraphics[width=0.85\textwidth]{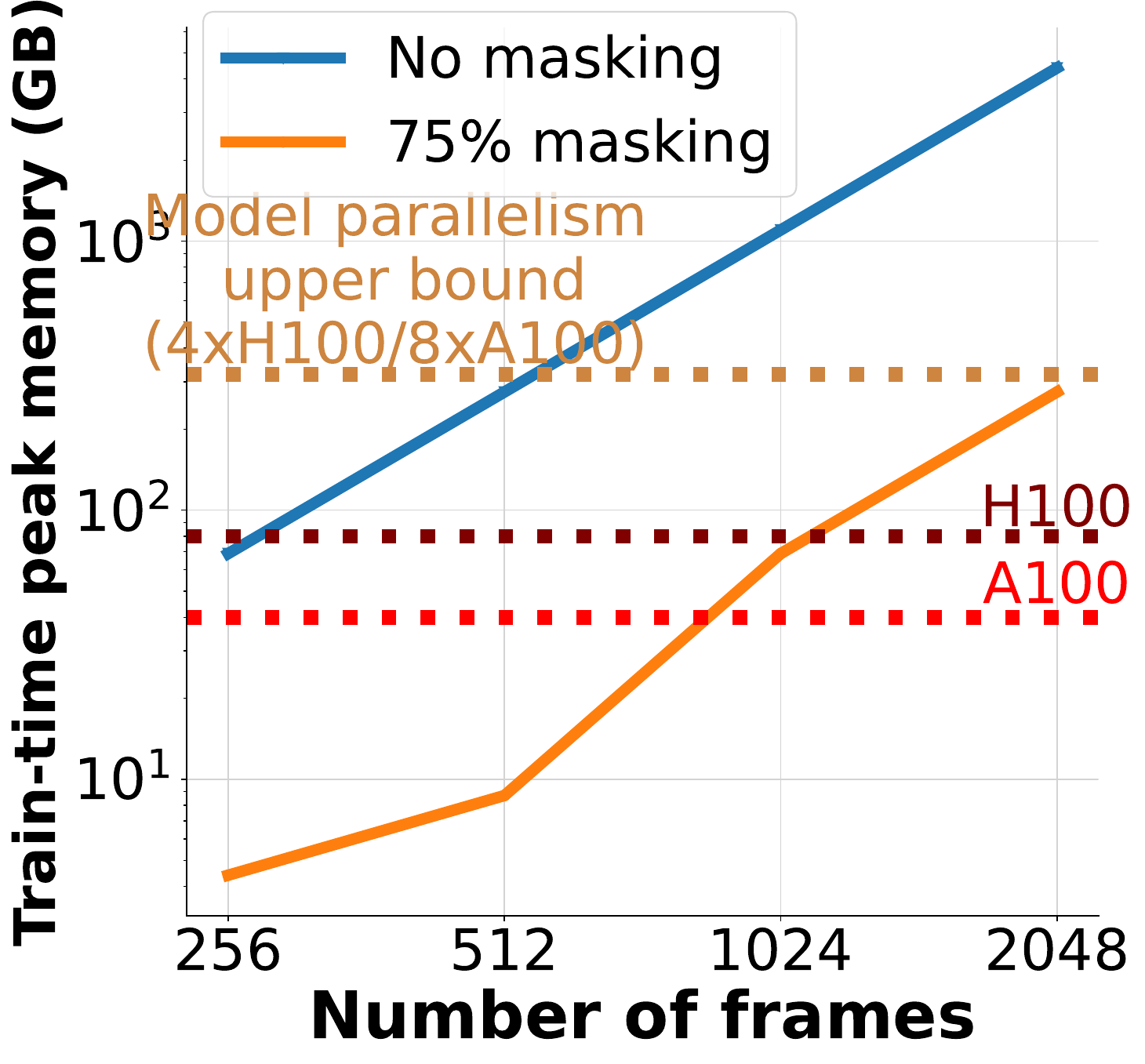}
        \label{fig:train_time_frames}
    \end{subfigure}
    \centering
    \hspace{3em}\begin{subfigure}[b]{0.3\textwidth}   
        \tiny
        \centering
        \includegraphics[width=0.85\textwidth]{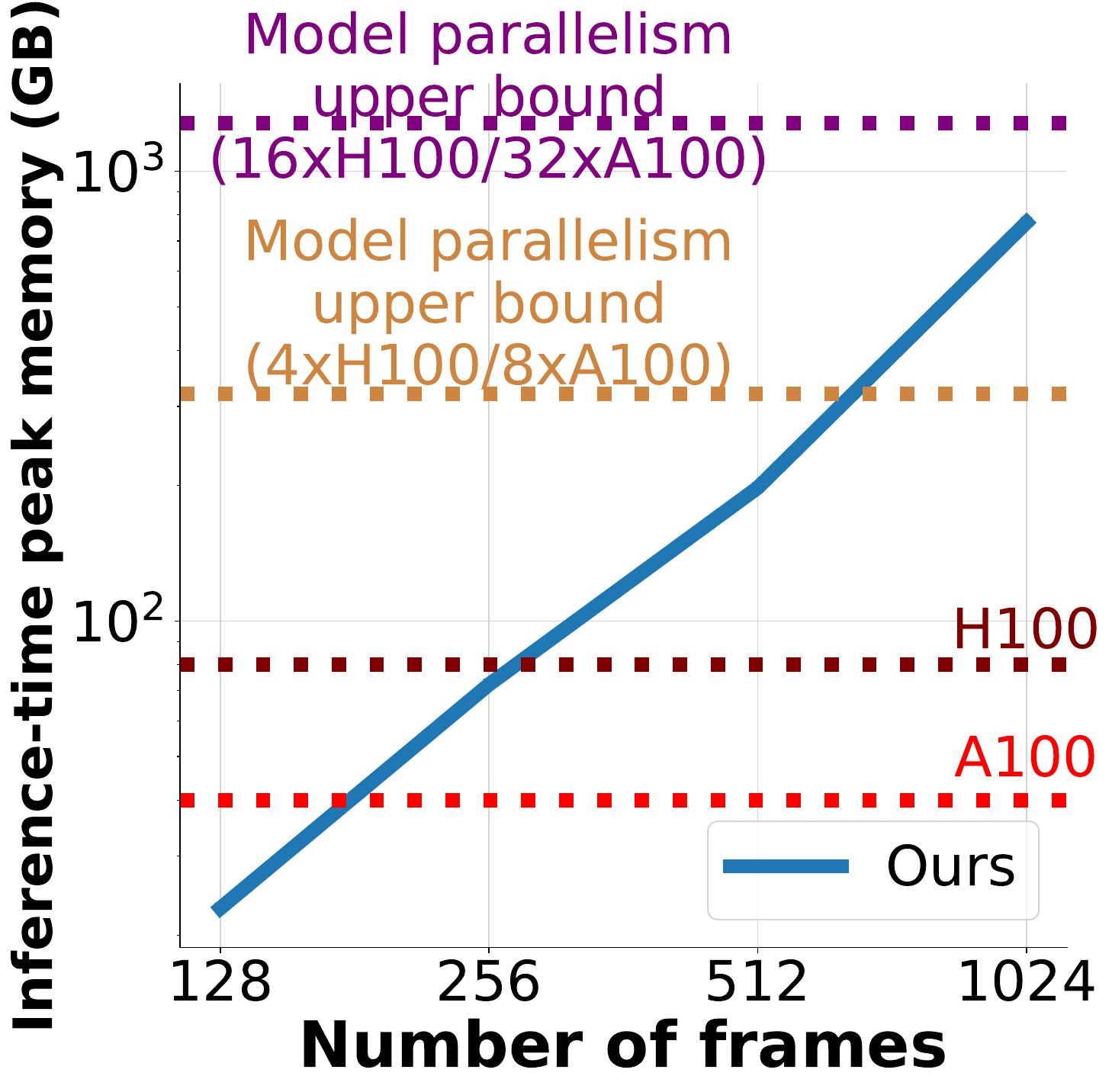}
        \label{fig:inference_time_frames}
    \end{subfigure}

    \caption[]
    {\footnotesize Peak train or inference time memory consumption with different model scales and number of input frames.}
    \label{fig:appendix_mem_figure}
\end{figure*}

\section{Memory Consumption and Scaling} \label{sec:appendix_memory}

We present the peak memory consumption of our model versus a vanilla joint space-time video encoder when scaling in model size or input length in Figure~\ref{fig:appendix_mem_figure}. First, we can scale the size of our video encoder up to 6B parameters without requiring model sharding when using A100 GPUs during training (first panel of Figure~\ref{fig:appendix_mem_figure}). Next, we can further scale the input sequence length to 2048 frames during training time when applying 75\% token masking and to 1024 frames during inference considering a smaller percentage of masking (i.e.,~30\% for which we have noticed no significant performance drop) when sharding the model parameters across 8 A100 GPUs (second and third panels of Figure~\ref{fig:appendix_mem_figure}).

\begin{table*}[t]
\footnotesize
\centering
\begin{tabular}{@{}lL{50em}@{}}
\toprule
\multicolumn{2}{c}{\includegraphics[width=0.95\textwidth]{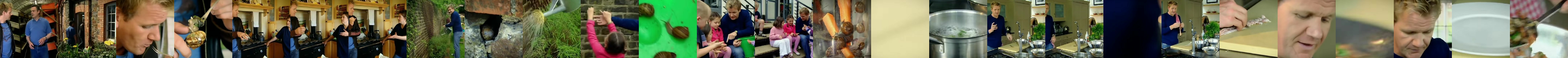}} \\
PALI-to-Bard & \textcolor{red}{A man in a blue shirt is talking to another man. The man in the blue shirt is holding a piece of paper.} They are talking about snails. The man in the blue shirt is holding a bowl of \textcolor{ForestGreen}{snails}. He is putting the \textcolor{ForestGreen}{snails} in a glass jar filled with water. \textcolor{red}{He is cooking in a kitchen. He is picking herbs from a bunch of green plants. He is cooking food on a gas stove.} He is cutting up a bunch of green vegetables on a white plate. He is putting a salad on a plate with a spoon. \\ \midrule
\longvivit & \textcolor{ForestGreen}{This is a video of cooking snails.} They first use a knife to cut the \textcolor{ForestGreen}{snails}. Then they put the \textcolor{ForestGreen}{snails} in a pot and cook them. After that they put the \textcolor{ForestGreen}{snails} in a bowl and put the sauce on them. Finally they eat the \textcolor{ForestGreen}{snails}. \\
\toprule
\multicolumn{2}{c}{\includegraphics[width=0.95\textwidth]{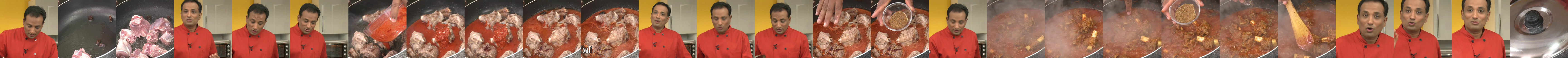}} \\
PALI-to-Bard & \textcolor{ForestGreen}{The video shows a chef preparing a dish called Van Rani Vab. The chef starts by mixing spices in a bowl. He then adds the spices to a pot of meat and sauce. He cooks the meat and sauce on the stove, stirring occasionally. Once the meat is cooked, the chef adds tomato sauce and saffron to the pot. He then cooks the dish for a few more minutes. Finally, the chef plates the dish and garnishes it with a mint leaf.} \textcolor{red}{The chef wears a red chef's jacket throughout the video. He is seen standing in front of a yellow wall, a microwave oven, and an oven. He also uses a variety of kitchen utensils, including a spoon, a fork, and a wooden spoon.}  The video ends with the chef eating the dish. He is seen holding a plate of food in front of a sign that says "Van". \\ \midrule
\longvivit & This is a video of cooking \textcolor{ForestGreen}{lamb shish kebab}. They first cut the \textcolor{ForestGreen}{lamb into small pieces}. Then they put the \textcolor{ForestGreen}{lamb in a pot} and add the spices. After that they add the onions and cook it. Finally they add the water and cook it. \\
\toprule
\multicolumn{2}{c}{\includegraphics[width=0.95\textwidth]{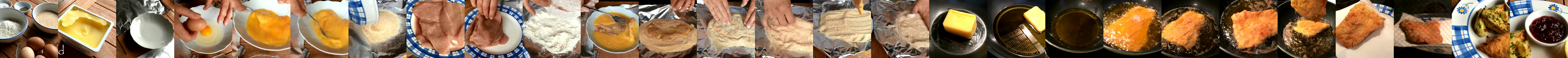}} \\
PALI-to-Bard & \textcolor{ForestGreen}{The video shows a person making Wiener Schnitzel.} First, the person cracks an egg into a bowl and mixes it with a fork. Then, they pour flour on a piece of tin foil and coat a piece of meat in the flour. Next, they dip the meat in the egg mixture and then coat it in bread crumbs. Finally, they cook the meat in a frying pan and serve it with potatoes and sauce. \\ \midrule
\longvivit & This is a video of making \textcolor{ForestGreen}{fried chicken}. \textcolor{red}{They first put the chicken into the frying pan and cook it for four minutes. Then they add the flour, egg, salt, pepper, and the mayonnaise. After that they fry the chicken for four minutes.} Finally they put the chicken on a plate and add the mayonnaise. \\
\bottomrule
\end{tabular}
\caption{Generated summaries of full-length YouCook2 videos by PALI-to-Bard and our~\longvivit model.}
\label{tab:appendix_examples}
\end{table*}

\section{Examples} \label{sec:appendix_examples}

Finally, we present examples of the generated video summaries for the full length videos of YouCook2 in Table~\ref{tab:appendix_examples}. For each example, we provide sparsely sampled frames from the entire video and the generated summaries by PALI-to-Bard, which is the most competitive model using LLM-assistance for information aggregation, and our~\longvivit model. As observed, PALI-to-Bard is able to provide very detailed descriptions of parts of the video, but it lacks consistency and coherence between the sentences of the output. Moreover, the model cannot focus on what is important in the video. For example, the main goal of the first video displayed in Table~\ref{tab:appendix_examples} is to demonstrate how to cook snails. However, the video includes several non-important clips, where people are discussing and/or change places from inside to outside and inside again. As a result, the summary provided by the model lacks coherence, describes isolated events and does not focus enough on the main point and ingredient of the video, which is the snails. In contrast, our model, although much smaller, is able to process the entire input sequence in one go and therefore provides more concise summaries, which remain coherent and do not include unimportant events. However, our model is still small-scale and does not employ an LLM for generation, which might lead to mistakes in the output. For example, the last video of Table~\ref{tab:appendix_examples} displays how to cook fried Schnitzel. In the generated summary by~\longvivit the frying of the chicken is repeated before and after adding the other ingredients (e.g., flour, salt, pepper), which is not true according to the video. We hypothesize that such mistakes can be eliminated when scaling the model size, and especially the LM component of our model.


\end{document}